\documentclass{article}

\usepackage{neurips_data_2024}

\usepackage[utf8]{inputenc} 
\usepackage[T1]{fontenc}    
\usepackage{hyperref}       
\usepackage{url}            
\usepackage{booktabs}       
\usepackage{amsfonts}       
\usepackage{nicefrac}       
\usepackage{microtype}      
\usepackage{xcolor}         
\usepackage{multirow}
\usepackage{subfigure}
\usepackage{graphicx}
\usepackage{wasysym}
\usepackage{floatrow}
\newfloatcommand{capbtabbox}{table}[][\FBwidth]
\usepackage{wrapfig}
\usepackage{enumitem}
\usepackage{CJK}

\usepackage{lineno}

\definecolor{ForestGreen}{RGB}{34, 139, 34}

\newcommand{\exprimentfontsize}{\fontsize{8}{11}\selectfont} 
\newcommand{\TaskAblationfontsize}{\fontsize{9}{12}\selectfont} 
\newcommand{\tablefontsize}{\fontsize{6}{9}\selectfont} 

\title{Emotion and Intent Joint Understanding in Multimodal Conversation: A Benchmarking Dataset}

\author{%
  Rui Liu\\
  Inner Mongolia University\\
  \texttt{liurui\_imu@163.com} \\
  \And
  Haolin Zuo \\
  Inner Mongolia University \\
  \texttt{zuohaolin\_0613@163.com} \\
  \AND
  Zheng Lian \\
  Chinese Academy of Sciences \\
  \texttt{lianzheng2016@ia.ac.cn} \\
  \And
  Xiaofen Xing \\
  South China University of Technology \\
  \texttt{xfxing@scut.edu.cn} \\
  \And
  Bj{\"o}rn W.\ Schuller \\
  Technical University of Munich \\
  \texttt{bjoern.schuller@imperial.ac.uk} \\
  \And
  Haizhou Li \\
  The Chinese University of Hong Kong, Shenzhen \\
  \texttt{haizhouli@cuhk.edu.cn} \\
}

\begin{document}

\nolinenumbers
\maketitle

\begin{abstract}
Emotion and Intent Joint Understanding in Multimodal Conversation (MC-EIU) aims to decode the semantic information manifested in a multimodal conversational history, while inferring the emotions and intents simultaneously for the current utterance. MC-EIU is enabling technology for many human-computer interfaces. However, there is a lack of available datasets in terms of annotation, modality, language diversity, and accessibility. In this work, we propose an MC-EIU dataset, which features 7 emotion categories, 9 intent categories, 3 modalities, i.e., textual, acoustic, and visual content, and two languages, i.e., English and Mandarin. Furthermore, it is completely open-source for free access. To our knowledge, MC-EIU is the first comprehensive and rich emotion and intent joint understanding dataset for multimodal conversation. Together with the release of the dataset, we also develop an Emotion and Intent Interaction (EI$^2$) network as a reference system by modeling the deep correlation between emotion and intent in the multimodal conversation. With comparative experiments and ablation studies, we demonstrate the effectiveness of the proposed EI$^2$ method on the MC-EIU dataset. The dataset and codes will be made available at: https://github.com/MC-EIU/MC-EIU.
\end{abstract}

\section{Introduction}
\label{sec: Introduction}
\vspace{-2mm}

Emotion and Intent Joint Understanding in Multimodal Conversation (MC-EIU) aims to infer the emotional state and intent information simultaneously by modeling the semantic dependency among the multimodal conversation \cite{singh2022emoint}.
Unlike separate tasks of emotion or intent recognition \cite{lian2023mer,zuo2023exploiting,zhang2022mintrec}, the MC-EIU task provides richer information to assist machines in better understanding human needs and improve empathy in human-machine conversation \cite{poria2017review,deng2023cmcu}.
It holds significant potential for application in various human-computer interaction scenarios, including call center 
dialog systems \cite{danieli2015emotion}, conversational agents \cite{cowie2001emotion}, and mental health counseling \cite{ringeval2018avec}, etc.

The MC-EIU task is challenging and involves two key problems: 1) modeling multimodal contextual information, and 2) modeling the interaction between emotion and intent. To support such research, extensive work has been done by previous researchers in both dataset and methodology aspects.
1) \textbf{In terms of dataset}, multiple datasets have been proposed to annotate emotion or intent labels for human dialogue video clips. However, there is a lack of available datasets in terms of annotation, modality, language diversity, and accessibility. For example, the IEMOCAP \cite{busso2008iemocap}, MEISD \cite{firdaus2020meisd}, MELD \cite{poria2019meld}, and M3ED \cite{zhao2022m3ed} datasets annotate multimodal dialogue data with  5, 8, and 7 emotion categories respectively. However, they have limited annotation diversity as they only provide emotion labels and lack intent labels. The OSED \cite{welivita2020fine} is an open-source dataset that includes 32 emotions and 9 intents. However, it is limited in terms of modality and language diversity, as it only consists of English textual data.
 While EmoInt-MD \cite{singh2022emoint} aligns well with the MC-EIU task, it only provides English data, which lacks language diversity. Additionally, the provided open-source address of EmoInt-MD for accessing the data is not available.
2) \textbf{In terms of methodology}, some works utilize state-of-the-art network architectures to build multi-task frameworks for joint recognition of emotion and intent. Specifically, the EmoInt-Trans \cite{singh2022emoint} just utilizes the embedding of neighboring utterances to boost the representation for the current utterance, and then adopts the Transformer with two projection heads to predict the emotion and intent labels. This approach neglects the complex dependency relationships within the multimodal dialog history and the deep interaction between emotion and intent information, which have been proven to be crucial in multimodal dialogue understanding \cite{deng2023cmcu}.

In this work, we propose the MC-EIU dataset that simultaneously fulfills four attributes, namely annotation, modality, language diversity, and accessibility, to support the research on the MC-EIU task.
This dataset consists of 4,970 conversational video clips from 3 English and 4 Chinese TV series, offering dialog scenarios closely related to the real world. It comprises annotations for 7 emotions and 9 intents, encompassing textual, acoustic, and visual modalities, in a total of 45,009 English utterances and 11,003 Mandarin utterances. Importantly, it is fully open-source, allowing for accessibility.
To the best of our knowledge, this dataset is the first comprehensive and rich emotion and intent joint understanding dataset for multimodal conversation, which can promote research of affective computing for the English and Chinese languages.

Additionally, we also develop an Emotion and Intent Interaction (EI$^2$) framework as a reference system by modeling the crucial deep correlation between emotion and intent in the multimodal conversation while considering the multimodal dialog context.
Specifically, We design a \textit{multimodal history encoder} to model the conversation history adequately. Then, we thoroughly integrate the conversation history features with the emotion and intent features of the current utterance. Finally, we propose an \textit{emotion-intent interaction encoder} to learn the complex interaction correlations between emotion and intent.
In summary, the main contributions of this work are as follows:
\begin{itemize}
\vspace{-1mm}
    \item We construct the first comprehensive and rich multimodal conversational emotion and intent joint understanding dataset, termed MC-EIU, which fulfills four attributes: annotation, modality, language diversity, and accessibility. It can promote research on affective computing and human-computer interaction. 
    \vspace{-0.5mm}
    \item We further propose the accompanying system, that is the Emotion and Intent Interaction (EI$^2$) framework, for the MC-EIU task. It involves modeling multimodal conversational history and deep emotion-intent interaction.
    \vspace{-0.5mm}
    \item The comparative experimental results demonstrate that our EI$^2$ surpasses all advanced baselines on our benchmarking dataset. Ablation and case studies further validate our EI$^2$.
    
\end{itemize}

\section{Related Work}
\label{sec: Related Work}
\vspace{-2mm}

\subsection{Related Dataset}
\label{Related Dataset}
\vspace{-2mm}

Table \ref{tab: Comparison with other datasets} provides an overview and summary of some of the most important datasets related to our work.
To highlight the significance of our MC-EIU dataset, we conduct a detailed comparison of these datasets based on annotation, modality, language diversity, and accessibility.

\textbf{Annotation Diversity}: The EmpatheticDialogues \cite{rashkin2019towards}, EmotionLines \cite{chen2018emotionlines}, DailyDialog \cite{li2017dailydialog}, EmoV-DB \cite{adigwe2018emotional}, IEMOCAP, MELD, MEISD, and M3ED are widely used for emotion recognition, but these datasets primarily focus on emotion labels and lack annotation diversity as they do not include intent labels.
Similarly, the Behance Intent Discovery dataset \cite{maharana2022multimodal} and MIntRec \cite{zhang2022mintrec} are limited in annotation diversity because they are only used for intent recognition and do not provide emotion labels.
In comparison, the Twitter Customer Support\cite{herzig2016classifying}, ED \cite{welivita2020taxonomy}, OSED, and EmoInt-MD contain the advantage of including both emotion and intent labels.

\textbf{Modality Diversity}: The EmpatheticDialogues, Emotionlines, DailyDialog, Twitter Customer Support \cite{herzig2016classifying}, ED \cite{welivita2020taxonomy}, and OSED offer a vast collection of conversational textual data.
The EmoV-DB is built with speech data in five emotion categories.
However, these datasets only provide information on a single modality and do not include multiple modalities simultaneously. Consequently, the absence of support for multiple modalities limits the modality diversity of these datasets.
Different from them, the IEMOCAP, MELD, MEISD, M3ED, Behance Intent Discovery dataset \cite{maharana2022multimodal}, MIntRec \cite{zhang2022mintrec}, and EmoInt-MD offer a broader range of modalities, including textual, acoustic, and visual data. 

\textbf{Language Diversity}: Most datasets in Table \ref{tab: Comparison with other datasets} mainly comprise English data, while the M3ED dataset solely consists of Mandarin data. Consequently, these datasets exhibit limitations in terms of language diversity.
Only EmoV-DB stands out as it includes both English and French data.  

\textbf{Accessibility}: All the datasets listed in Table \ref{tab: Comparison with other datasets} are accessible except for MEISD and Twitter. It is worth mentioning that the dataset from the provided open-source address in EmoInt-MD is currently unavailable, which poses a challenge in accessing the data.

In a nutshell, there is a lack of available data on the aforementioned four aspects of the MC-EIU task. Our dataset addresses these gaps and provides data that fulfill all four aspects.
\vspace{-2mm}

\newcommand{\datasetfontsize}{\fontsize{6}{8}\selectfont} 

\setlength{\tabcolsep}{0.01 \linewidth}{
\begin{table*}
    \datasetfontsize
    \centering
\begin{tabular}{l|c|c|c|c|c|c|c|c|c|c}
\hline
\textbf{Datasets}                                                                                        & \textbf{Emos} & \textbf{Ints} & \textbf{Modality} & \textbf{Language}                                             & \textbf{\begin{tabular}[c]{@{}c@{}}Dias\\ (k)\end{tabular}} & \textbf{\begin{tabular}[c]{@{}c@{}}Utts\\ (k)\end{tabular}} & \textbf{AD} & \multicolumn{1}{l|}{\textbf{MD}} & \multicolumn{1}{l|}{\textbf{LD}} & \textbf{AC}    \\ \hline
EmpatheticDialogues \cite{rashkin2019towards}                 & \checked         & -             & T                   & English                                                       & 24.8                                                        & 107.2                                                       & \textcolor{red}{$\times$}                                               & \textcolor{red}{$\times$}                                                                  & \textcolor{red}{$\times$}                                                                  & \textcolor{black}{\textcircled{\textcolor{white}{\textbf{\footnotesize$\emptyset$}}}}                    \\ \hline
Emotionlines \cite{chen2018emotionlines}                                                                 & \checked         & -             & T                   & English                                                       & 1.0                                                         & 14.5                                                        & \textcolor{red}{$\times$}                                               & \textcolor{red}{$\times$}                                                                  & \textcolor{red}{$\times$}                                                                  & \textcolor{black}{\textcircled{\textcolor{white}{\textbf{\footnotesize$\emptyset$}}}}                    \\ \hline
DailyDialog \cite{li2017dailydialog}                                                                     & \checked         & -             & T                   & English                                                       & 12.2                                                        & 103.6                                                       & \textcolor{red}{$\times$}                                               & \textcolor{red}{$\times$}                                                                  & \textcolor{red}{$\times$}                                                                  & \textcolor{black}{\textcircled{\textcolor{white}{\textbf{\footnotesize$\emptyset$}}}}                    \\ \hline
EmoV-DB \cite{adigwe2018emotional}                                                                       & \checked         & -                                  & A                   & \begin{tabular}[c]{@{}c@{}}English +  French\end{tabular}   & -                                                         & 7.6                                                       & \textcolor{red}{$\times$}                                                         & \textcolor{red}{$\times$}                                                                  & \textcolor{black}{\textcircled{\textcolor{white}{\textbf{\footnotesize$\emptyset$}}}}                                                                                    & \textcolor{black}{\textcircled{\textcolor{white}{\textbf{\footnotesize$\emptyset$}}}}                                                                                   \\ \hline
IEMOCAP \cite{busso2008iemocap}                                                                          & \checked         & -             & T+A+V               & English                                                       & 0.2                                                         & 7.4                                                         & \textcolor{red}{$\times$}                                               & \textcolor{black}{\textcircled{\textcolor{white}{\textbf{\footnotesize$\emptyset$}}}}                                                                                     & \textcolor{red}{$\times$}                                                                  & \textcolor{black}{\textcircled{\textcolor{white}{\textbf{\footnotesize$\emptyset$}}}}                    \\ \hline
MELD \cite{poria2019meld}                                                                                & \checked         & -             & T+A+V               & English                                                       & 1.4                                                         & 13.7                                                        & \textcolor{red}{$\times$}                                               & \textcolor{black}{\textcircled{\textcolor{white}{\textbf{\footnotesize$\emptyset$}}}}                                                                                     & \textcolor{red}{$\times$}                                                                  & \textcolor{black}{\textcircled{\textcolor{white}{\textbf{\footnotesize$\emptyset$}}}}                    \\ \hline
MEISD \cite{firdaus2020meisd}                                                                             & \checked         & -             & T+A+V               & English                                                       & 1.0                                                         & 20.0                                                        & \textcolor{red}{$\times$}                                               & \textcolor{black}{\textcircled{\textcolor{white}{\textbf{\footnotesize$\emptyset$}}}}                                                                                     & \textcolor{red}{$\times$}                                                                  & \textcolor{red}{$\times$} \\ \hline
M3ED \cite{zhao2022m3ed}                                                                                  & \checked         & -             & T+A+V               & Mandarin                                                      & 1.0                                                         & 24.4                                                        & \textcolor{red}{$\times$}                                               & \textcolor{black}{\textcircled{\textcolor{white}{\textbf{\footnotesize$\emptyset$}}}}                                                                                     & \textcolor{red}{$\times$}                                                                  & \textcolor{black}{\textcircled{\textcolor{white}{\textbf{\footnotesize$\emptyset$}}}}                    \\ \hline \hline
Behance Intent Discovery dataset \cite{maharana2022multimodal} & -             & \checked         & T+A+V               & English                                                       & -                                                          & 20.0                                                        & \textcolor{red}{$\times$}                                               & \textcolor{black}{\textcircled{\textcolor{white}{\textbf{\footnotesize$\emptyset$}}}}                                                                                     & \textcolor{red}{$\times$}                                                                  & \textcolor{black}{\textcircled{\textcolor{white}{\textbf{\footnotesize$\emptyset$}}}}                    \\ \hline
MIntRec \cite{zhang2022mintrec}                                                                          & -             & \checked         & T+A+V               & English                                                       & -                                                          & 2.2                                                         & \textcolor{red}{$\times$}                                               & \textcolor{black}{\textcircled{\textcolor{white}{\textbf{\footnotesize$\emptyset$}}}}                                                                                     & \textcolor{red}{$\times$}                                                                  & \textcolor{black}{\textcircled{\textcolor{white}{\textbf{\footnotesize$\emptyset$}}}}                    \\ \hline \hline
Twitter Customer Support \cite{herzig2016classifying}          & \checked         & \checked         & T                   & English                                                       & 2.4                                                         & 14.1                                                        & \textcolor{black}{\textcircled{\textcolor{white}{\textbf{\footnotesize$\emptyset$}}}}                                                                  & \textcolor{red}{$\times$}                                                                  & \textcolor{red}{$\times$}                                                                  & \textcolor{red}{$\times$} \\ \hline
ED \cite{welivita2020taxonomy}                                                                           & \checked         & \checked         & T                   & English                                                       & 24.8                                                        & 107.2                                                       & \textcolor{black}{\textcircled{\textcolor{white}{\textbf{\footnotesize$\emptyset$}}}}                                                                  & \textcolor{red}{$\times$}                                                                  & \textcolor{red}{$\times$}                                                                  & \textcolor{black}{\textcircled{\textcolor{white}{\textbf{\footnotesize$\emptyset$}}}}                    \\ \hline
OSED \cite{welivita2020fine}                                                                             & \checked         & \checked         & T                   & English                                                       & 1000.0                                                      & 3488.3                                                      & \textcolor{black}{\textcircled{\textcolor{white}{\textbf{\footnotesize$\emptyset$}}}}                                                                  & \textcolor{red}{$\times$}                                                                  & \textcolor{red}{$\times$}                                                                  & \textcolor{black}{\textcircled{\textcolor{white}{\textbf{\footnotesize$\emptyset$}}}}                    \\ \hline
EmoInt-MD \cite{singh2022emoint}                                                                         & \checked         & \checked         & T+A+V               & English                                                       & 32.4                                                        & 724.8                                                       & \textcolor{black}{\textcircled{\textcolor{white}{\textbf{\footnotesize$\emptyset$}}}}                                                                  & \textcolor{black}{\textcircled{\textcolor{white}{\textbf{\footnotesize$\emptyset$}}}}                                                                                     & \textcolor{red}{$\times$}                                                                  & \textcolor{red}{$\times$} \\ \hline
\textbf{MC-EIU(Ours)}                                                                                    & \checked         & \checked         & T+A+V               & \begin{tabular}[c]{@{}c@{}}English + Mandarin\end{tabular} & 5.0                                                         & 56.0                                                        & \textcolor{green}{\textcircled{\textcolor{white}{\textbf{\footnotesize$\emptyset$}}}}                                                                  & \textcolor{green}{\textcircled{\textcolor{white}{\textbf{\footnotesize$\emptyset$}}}}                                                                                     & \textcolor{green}{\textcircled{\textcolor{white}{\textbf{\footnotesize$\emptyset$}}}}                                                                                     & \textcolor{green}{\textcircled{\textcolor{white}{\textbf{\footnotesize$\emptyset$}}}}                    \\ \hline
\end{tabular}
  \vspace{-2mm}
    \caption{Comparison with existing benchmark datasets. `T', `V', and `A' refer to textual, visual, and acoustic information. AD, MD, LD, AC denote Annotation Diversity, Modality Diversity, Language Diversity, and Accessibility, respectively}
      \vspace{-2mm}
    \label{tab: Comparison with other datasets}
\end{table*}
}

\subsection{Related Method}
\label{sec: Related Method}
\vspace{-2mm}

\textbf{Multi-task Prediction}: Multi-task learning aims to leverage the valuable information in multiple related tasks to enhance generalization performance across all tasks \cite{yang2017trace}, which has a wide range of applications in fields such as Emotion Recognition \cite{hazarika2020misa}, Object Detection \cite{li2023logonet}, Conversational Speech Synthesis \cite{liu2023emotion}, etc. Multi-task learning methods can be broadly classified into Hard Parameter Sharing (HPS) and Soft Parameter Sharing (SPS) \cite{zhang2022rethinking,pahari2022multi}.
HPS serves as the foundational deep neural network that is shared across different tasks, while each task utilizes its own distinct top layer \cite{zhang2022rethinking,hazarika2020misa}.
SPS allows the model to share some parameters between different tasks while preserving task-specific parameters \cite{pahari2022multi}. 

It should be noted that compared to HPS, the advantage of SPS lies in the ability of the network to utilize shared features among different tasks by sharing bottom-layer parameters. It allows for better utilization of interactive information between tasks \cite{pahari2022multi}.

\textbf{Emotion-Intent Interaction}: Previous works confirm that there is a strong interaction between emotion and intent \cite{welivita2020taxonomy,welivita2020fine,singh2022emoint,deng2023cmcu}.
For instance, \cite{singh2022emoint} emphasizes that the emotions of the speaker can be influenced by particular intents in dialogues.
To explore the interaction between emotion and intention, they built the EmoInt-Trans model that follows the HPS framework. It uses a MISA network \cite{hazarika2020misa} as the backbone, where the emotion and intent prediction tasks share all the parameters of this backbone.
However, the complex relationship between emotion and intent poses a challenge for HPS, as it may struggle to accommodate the distinct characteristics of each task.
Note that the biggest difference between our EI$^{2}$ model and EmoInt-Trans is that we employ an SPS mode to learn the deep-level interactive information for emotion and intent.
\vspace{-2mm}

\section{MC-EIU Dataset Construction}
\vspace{-2mm}

\subsection{Data Collection and Pre-processing}
\label{subacc: Data Collection and Pre-processing}
\vspace{-2mm}

\begin{wraptable}{r}{0.6 \textwidth}
\tablefontsize
\centering
\begin{tabular}{l|ccc|ccc}
\hline
\multirow{2}{*}{\textbf{Statistics}} & \multicolumn{3}{c|}{\textbf{English}}           & \multicolumn{3}{c}{\textbf{Mandarin}}           \\ \cline{2-7} 
                                     & \textbf{Train} & \textbf{Valid} & \textbf{Test} & \textbf{Train} & \textbf{Valid} & \textbf{Test} \\ \hline
\# Conversations                      & 2,807          & 400            & 806           & 667            & 95             & 195           \\ \hline
\# Utterances                         & 31,451         & 4,509          & 9,049         & 7,643          & 1,148          & 2,212         \\ \hline
\# Duration (hours)                   & 28.51          & 4.02           & 8.22          & 8.51           & 1.36           & 2.42          \\ \hline
Avg.\ UL                & 12.68          & 12.49          & 12.76         & 19.11          & 19.91          & 18.14         \\ \hline
Avg.\ \# of DU (seconds)               & 3.26           & 3.21           & 3.27          & 4.01           & 4.26           & 3.94          \\ \hline
Avg.\ \# of UC                         & 11.20          & 11.27          & 11.23         & 11.46          & 12.08          & 11.34         \\ \hline
Avg.\ \# of EC                         & 2.58           & 2.57           & 2.60          & 2.41           & 2.54           & 2.42          \\ \hline
Avg.\ \# of IC                         & 3.29           & 3.86           & 3.87          & 3.18           & 3.24           & 3.10          \\ \hline
\end{tabular}

\caption{Statistic of our MC-EIU. UL refers to the utterance length, DU denotes the duration per utterance, UC is the utterances per conversation, EC means the emotions per conversation, and IC means the intents per conversation.}

\label{tab: Statistic of MC-EIU}
\end{wraptable}

To simulate the emotional conversation scenarios in real-world situations \cite{zhao2022m3ed,singh2022emoint}, we select emotional video clips from 3 English (716 episodes) TV series and 4 Chinese (119 episodes) TV series across genres such as family, romance, crime, etc.\footnote{More details are shown in the Appendix.}
All videos are accompanied by corresponding subtitle files. Afterward, we preprocess the data to obtain the desired format and filter out low-quality data that does not meet the criteria. 
Specifically, 1) We first design Regular Expression scripts to extract text transcription and the timestamps from the subtitle; 
2) Then, we leverage VideoFileClip\footnote{\url{https://moviepy-tburrows13.readthedocs.io/en/improve-docs/ref/VideoClip/VideoFileClip.html}} to partition the videos into multiple clips based on the timestamps; 
3) At last, we engage crowd workers to pick out high-quality conversational segments from the same conversational scene.
Note that we provide the following specific guidelines to the crowd workers based on relevant work \cite{zhao2022m3ed}:
a) Conversational scenes should exclusively involve interaction between two speakers;
b) The selected conversational scenes should be free from noise or special effects sounds to ensure video quality is not affected;
c) Each conversational segment should encompass at least two rounds of interaction between the speakers to provide abundant context information;
d) The text content should be accurate and align with the video clips.
\vspace{-2mm}


\subsection{Data Annotation}
\label{subsec: Data Annotation}
\vspace{-2mm}
\begin{wrapfigure}{r}{0.55 \textwidth}
    \centering
    \subfigure[English]{
        \includegraphics[width=0.46\linewidth]{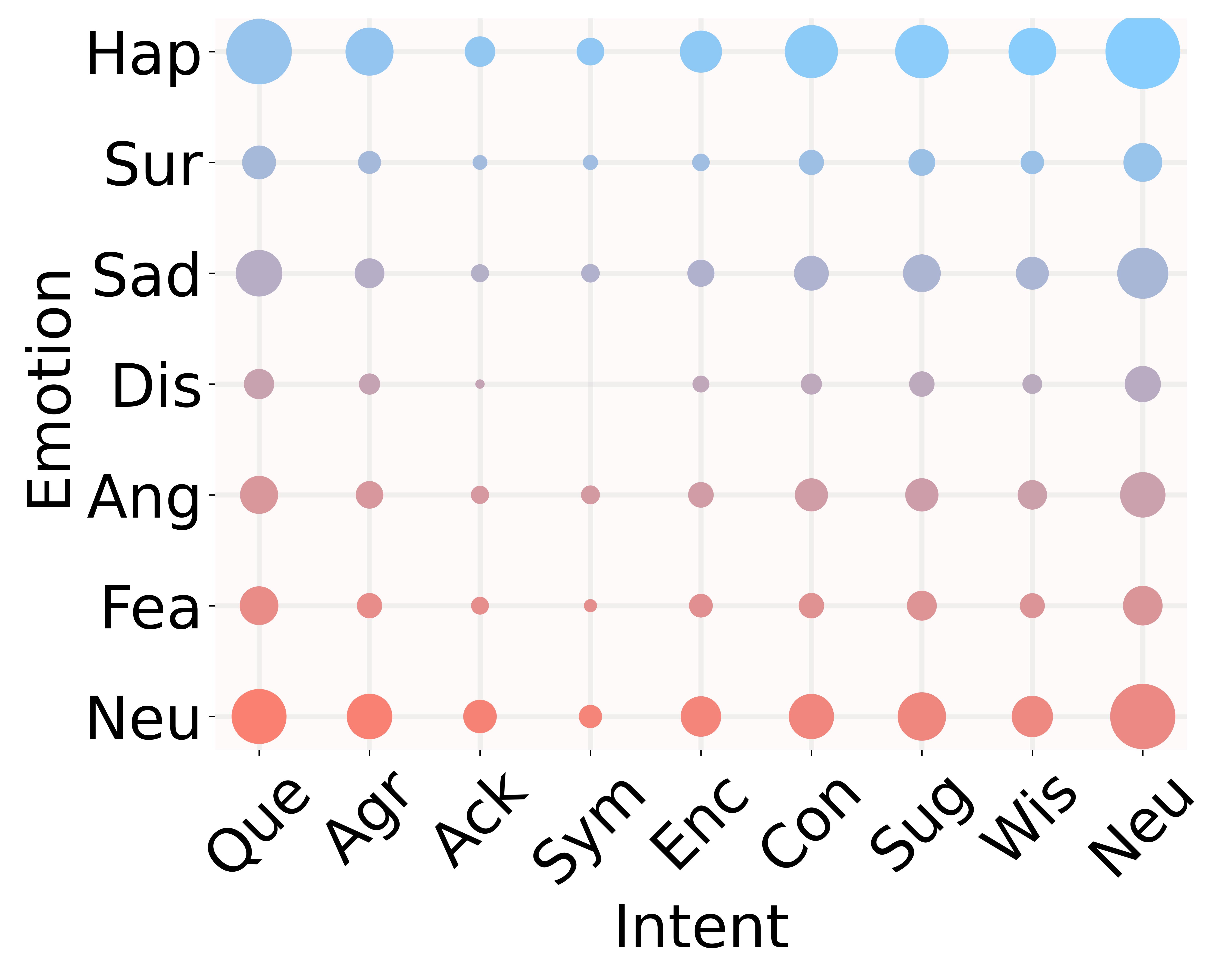}
    }
    \subfigure[Mandarin]{
        \includegraphics[width=0.46\linewidth]{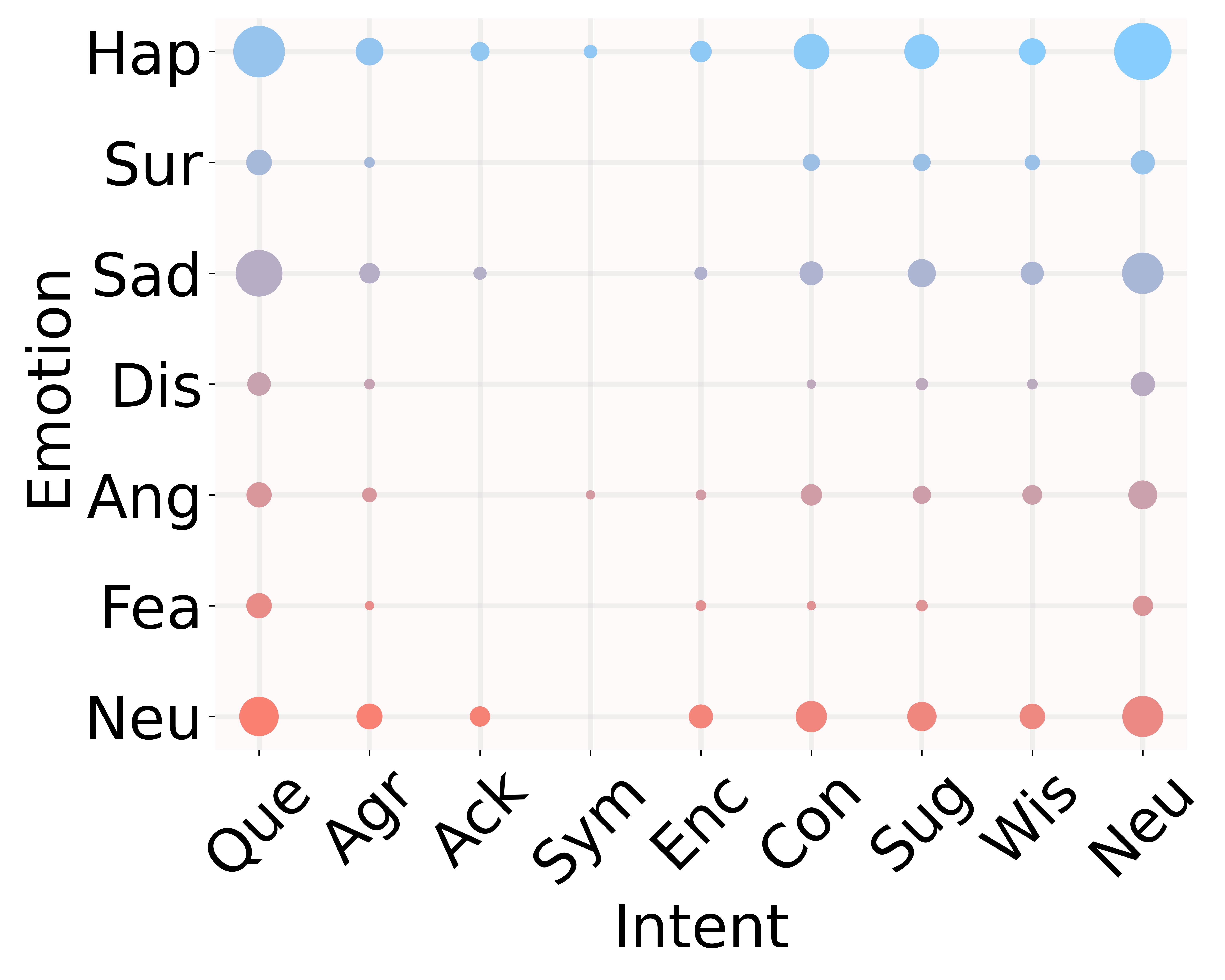}    
    }
    \caption{Visualization of the correlation between emotions and intents in the MC-EIU dataset.
    Each circle in the graph represents the sample count for a specific `emotion-intent' pair.
    }
    \label{fig: Correlation}
    \vspace{-2mm}
\end{wrapfigure}
\textbf{Annotation Scheme}: The annotation scheme consists of emotion, intent, and speaker annotation. 
For emotion annotation, we choose Ekman's six basic emotions (\textit{happy}, \textit{surprise}, \textit{sad}, \textit{disgust}, \textit{anger}, and \textit{fear}), along with the \textit{neutral} label, which is a 7-emotion annotation scheme widely used in previous works \cite{zhao2022m3ed,busso2008iemocap,poria2019meld}.
For intent annotation, we follow \cite{welivita2020taxonomy,welivita2020fine} and annotate each utterance based on nine intents: \textit{questioning}, \textit{agreeing}, \textit{acknowledging}, \textit{sympathizing}, \textit{encouraging}, \textit{consoling}, \textit{suggesting}, \textit{wishing}, and \textit{neutral}. 
For speaker annotation, we follow the scheme described in \cite{zhao2022m3ed} and assign the labels ``0'' and ``1'' to annotate the speaker for each dialog.
The above annotation scheme ensures consistency with real conversation scenarios \cite{zhao2022m3ed,poria2019meld}.

We recruit postgraduate students from the Foreign Languages College as annotators for our project\footnote{Although these annotators' native language is Chinese, they have passed the Test for English Majors Grade Eight (TEM-8), indicating their proficiency in English language skills, including listening, speaking, reading, and writing, comparable to native English speakers.}. Prior to commencing the annotation process, all annotators undergo a training and assessment phase. Only those who obtain a passing score are selected to proceed to the data annotation stage. Annotators who do not meet the passing criteria are required to undergo additional training and reassessment until they successfully pass the assessment.

\textbf{Annotation Process}: 
We recruited 21 annotators and divided them into seven groups. Each group consists of three volunteers, and non-duplicate data is assigned to each group for annotation. This ensures that each conversation data is annotated by three volunteers.
Each annotator can access the current utterance and history in the multimodal conversation, including text, audio, and video clips during the annotation process.
The annotators are asked to select the appropriate emotion and intent labels from the annotation scheme after watching the video clip. 
Simultaneously, annotators are required to assign speaker labels to each utterance in the dialog based on the speaking order.
Suppose annotators encounter difficulty in assigning a specific category to a video clip. In that case, they categorize the utterance as \textit{other}, and the corresponding dialog containing this utterance will not be included in our dataset.
\vspace{-2mm}

\subsection{Data Annotation Finalization}
\label{subsec: Data Annotation Finalization}
\vspace{-2mm}

\begin{wraptable}{l}{0.4 \textwidth}
    \centering
    \tablefontsize
    \begin{tabular}{l|cc}
        \hline
        \textbf{Dataset}      & \textbf{Emotion} & \textbf{Intent} \\ \hline
        IEMOCAP \cite{busso2008iemocap}               & 0.48  & -      \\
        MELD \cite{poria2019meld}                  &  0.43  & -      \\
        M3ED \cite{zhao2022m3ed}                  &  0.59  & -      \\
        OSED \cite{welivita2020fine}                  & 0.46  &  0.46 \\
        EmoInt-MD \cite{singh2022emoint}               & 0.69  & 0.57 \\ \hline
        \textbf{MC-EIU} \textit{(English)}        &  \textbf{0.57}  &  \textbf{0.59} \\
        \textbf{MC-EIU} \textit{(Mandarin)}      &  \textbf{0.54}  &  \textbf{0.57} \\ \hline
    \end{tabular}
  \vspace{-2mm}
\caption{Comparison of Fleiss's Kappa coefficients between other datasets and MC-EIU.}
\label{tab: Kappa}
\end{wraptable}
We employ a majority voting strategy on all utterance annotations to derive the final emotion and intent labels. If at least two annotators provide the same annotation for one utterance, it is considered the final label. If all three annotators have different annotations, an additional emotion expert is consulted to confirm the final label.


The statistics of the MC-EIU dataset are presented in Table \ref{tab: Statistic of MC-EIU}.
It comprises a total of 56,012 utterances, including 4,013 conversations in English and 957 conversations in Mandarin. 
The total duration of the MC-EIU set is 53.06 hours.
We partition the dataset into the train, valid, and test sets with a ratio of 7:1:2 as stated in \cite{zhao2022m3ed,poria2019meld}.
To assess the reliability of our data annotation, we compute Fleiss's Kappa ($\kappa$) \cite{fleiss2013statistical} for emotion and intent annotation independently, as presented in Table \ref{tab: Kappa}.
Note that $\kappa$ of our dataset is higher than or comparable to that of previous studies, which serves as strong evidence for the reliability and accuracy of our annotation. 

We further explore the correlation between emotion and intent.
We created two 7x9 two-dimensional matrices, with each element representing the number of samples in the dataset for each ``emotion-intent'' pair.
Using the sample count as the radius, we visualize circles at the corresponding matrix positions. The resulting visualization is depicted in Figure \ref{fig: Correlation}.
Larger circles indicate more samples and higher correlation.
It can be observed that emotions and intents are not strictly one-to-one correspondences, and different intents vary in their influence on specific emotions, and vice versa. 
Take Figure.\ref{fig: Correlation} (a) as an example, compared to ``Hap-Sym", ``Hap-Agr" occurs at a higher frequency, indicating that ``Agreeing" is more likely to drive the expression of ``Happy". Similarly, compared to ``Dis-Que", ``Sad-Que" has a higher occurrence rate, suggesting that ``Questioning" is more closely associated with ``Sad" compared to ``Disgust".
Additionally, we observe that the correlation between emotion and intent in the English dataset is more intricate compared to the Mandarin dataset. For example, the emotion "Sur" is associated with all intent categories in the English dataset, whereas in the Mandarin dataset, it is only linked to 6 intent categories ("Que", "Agr", "Con", "Sug", "Wis", and "Neu").
Such a complex relationship introduces a significant challenge to the MC-EIU task.
We make more details of data annotation finalization in the Appendix.
\vspace{-2mm}

\begin{figure*}[!t]
    \centering
    \includegraphics[width=0.95\linewidth]{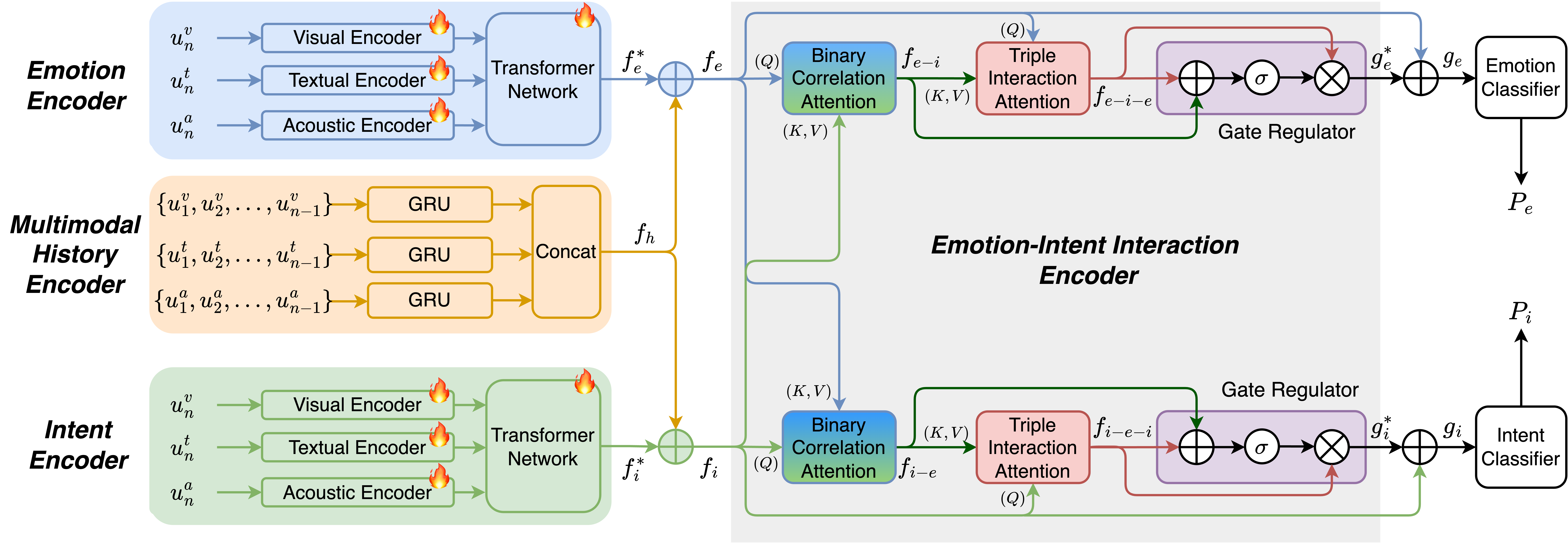}
    \caption{Overview of the Emotion-Intent Interaction (EI$^{2}$) Network. The modules with fire symbols indicate the need for pre-training.}
    \label{fig: The overall architecture of the MC-EIU framework}
    \vspace{-2mm}
\end{figure*}

\section{Emotion-Intent Interaction (EI$^2$) Network}
\label{sec: Method}
\vspace{-2mm}

\subsection{Overall Architecture}
\vspace{-2mm}

To model the multimodal dialog history and the deep-level interaction between emotion and intent, we propose an EI$^{2}$ network as shown in Figure \ref{fig: The overall architecture of the MC-EIU framework}.
The network consists of the following components: 1) \textit{Emotion \& Intent Encoders} aim to generate the multimodal emotion and intent representations for current utterance; 2) the \textit{Multimodal History Encoder} is responsible for capturing the multimodal contextual semantic information from the multimodal history; 3) the \textit{Emotion-Intent Interaction Encoder} is proposed to learn the deep interaction between emotions and intents in a conversation; 4) \textit{Emotion \& Intent Classifiers} aim to make predictions based on the emotion-intent interactive information.

\textbf{Emotion \& Intent Encoders}: In contrast to previous work \cite{singh2022emoint} that extracts general semantic representations from utterances, we add separate feature encoders for emotion and intent understanding to extract more explicit emotion and intent features. 

The emotion and intent encoders share a similar structure, which includes a Visual Encoder, a Textual Encoder, an Acoustic Encoder, and a Transformer fusion network.
Assume that $u_n^a$, $u_n^t$, $u_n^v$ represent the acoustic, textual, and visual features of the $n$-th utterance $u_n$, the multimodal emotion representation $f^*_e$ and intent representation $f^*_i$ can be expressed as:
\begin{equation}
    f^*_s = F^*_s(Concat(F^v_s(u^v_n), F^a_s(u^a_n), F^t_s(u^t_n))),
    \label{eq: emotion and intent information of n-th utterance}
\end{equation}
where $s \in \{e, i\}$, $F^a_s$ is the Acoustic Encoder based on LSTM \cite{sak2014long} and max-pooling, $F^v_s$ is the Visual Encoder that has the same structure as $F^a_s$, $F^t_s$ is the TextCNN-based Text Encoder \cite{kim2014convolutional} adopted, and the $F^*_s$ means the Transformer fusion network.

It is worth noting that the modules with fire symbols in Figure 2 indicate the need for pre-training our intent and emotion encoders. Further details will be provided in Section \ref{subsec: Training Strategy}.

\textbf{Multimodal History Encoder}: 
Different from EmoInt-Trans \cite{singh2022emoint}, which solely models the context information of adjacent utterances, our Multimodal History Encoder takes into account a broader range of historical information.
For the current utterance $u_n$, the multimodal conversational history information $f_h$ can be represented as:
\begin{equation}
    f^m_h = F_h^m(\{u^m_1,u^m_2,...,u^m_{n-1}\})
\end{equation}


\begin{equation}
    f_h = Concat(f^v_h, f^a_h, f^t_h)
\end{equation}
where $m \in \{v, a, t\}$, $F_h^m$ denotes the GRU-based History Encoder \cite{lee2023dailytalk}.
Subsequently, $f_h$ is independently fused with $f_e$ and $f_i$ to obtain the robust emotion feature $f_e$ and intent feature $f_i$: $ f_s = f^*_s + f_h$.



\textbf{Emotion-Intent Interaction Encoder}: 
The approach of simply sharing the hidden state is not sufficient to achieve explicit information transfer between two tasks. Therefore, we propose the \textit{Emotion-Intent Interaction Encoder} to learn the deep interactive information between them by considering their complex correlations. As shown in Figure \ref{fig: The overall architecture of the MC-EIU framework}, the Emotion-Intent Interaction Encoder contains two branches for emotion or intent prediction, where each branch consists of Binary Correlation Attention, Triple Interaction Attention, and Gate Regulator.

\textbf{Binary Correlation Attention} first employs cross-attention to learn the mutual influence between emotions and intentions, which can also be referred to as the binary correlation between the two tasks. Specifically, we first adopt linear projection to map the $f_e$ and $f_i$ to generate the corresponding $Q$, $K$, and $V$.
Then, the attention mechanism is used to extract the correlation between $f_e$ and $f_i$:
\begin{equation}
    f_{\gamma-\beta} = Attention(f^*_{\gamma}, f^*_{\beta}, f^*_{\beta}),
\end{equation}
where $\gamma, \beta \in \{e, i\}$, if $\gamma$ denotes $e$, then, $\beta$ denotes $i$, and vice versa.

\textbf{Triple Interaction Attention} further explores the deep interactive information between the two tasks by integrating the binary correlation and task-specific information from each branch to obtain a more comprehensive and in-depth task interaction feature. Inspired by \cite{jiang2023empathy}, we propose a triple interaction attention mechanism to compute cascaded interaction feature representations.
It takes the outputs of binary correlation attention, along with the $f_e$ and $f_i$, as inputs; the deep interactive information $f_{\gamma-\beta-\gamma}$ of two tasks can be obtained by:
\begin{equation}
    f_{\gamma-\beta-\gamma} = Attention(f^*_{\gamma}, f^*_{\gamma-\beta}, f^*_{\gamma-\beta})
\end{equation}

\textbf{Gate Regulator:} It utilizes a gating mechanism to automatically learn the weights of the binary correlation features and ternary interaction features, to model the potential impact of the correlation between different emotion-intent pairs in Figure \ref{fig: Correlation} on the interaction features. This allows us to adjust the contribution of the binary correlation to the final recognition of emotions or intents.
Specifically, the gate regulator first adds $f_{\gamma-\beta-\gamma}$ and $f_{\gamma-\beta}$ together. Afterward, the Sigmoid function is applied to obtain a control gate value between them. Lastly, $f_{\gamma-\beta}$ is multiplied by the control gate value to effectively adjust the weight of the correlation information:
\begin{equation}
    g^*_{\gamma} = f_{\gamma-\beta-\gamma} * sigmoid(f_{\gamma-\beta-\gamma} + f_{\gamma-\beta}),
\end{equation}
where $g^*_{\gamma}$ represents the final output of Emotion-Intent Interaction Encoder.
With the collaboration of these three components, our EI$^2$ model captures deep-level interaction between emotion and intent.

\textbf{Emotion \& Intent Classifiers: }
To preserve the specific information of the emotion and intent representations while incorporating the deep interactive information, we perform the residual connection to combine the $g^*$ with the emotion feature $f_e$ and the intent feature $f_i$ separately before making the final prediction.
At last, the Emotion and Intent Classifiers take the results of the residual connection to predict the final emotion and intent categories, respectively:
\vspace{-1mm}
\begin{equation}
    P_e = CLS_e(g_{e} = g^*_{e} + f_{e}); P_i = CLS_i(g_{i} = g^*_{i} + f_{i}).
\end{equation}

\subsection{Training Strategy}
\label{subsec: Training Strategy}
\vspace{-2mm}

We first pre-train the emotion and intent encoders to ensure the effective extraction of emotion and intent information.
given the presence of category imbalance in the dataset (shown in the Appendix), we employ the Focal Loss (FL) \cite{lin2017focal} as the loss function $\mathcal{L}_{pre}$ in the pre-training phase to constrain the prediction of the emotion and intent to be close to the Ground Truth $\hat{P}_e$ and $\hat{P}_i$ of emotion and intent and to improve the model's ability to focus on the categories of a small sample:
\begin{equation}
    P^*_e = CLS^*_e(f^*_e); \quad P^*_i = CLS^*_i(f^*_i)
\end{equation}
\begin{equation}
    \mathcal{L}_{pre} = FL(\hat{P}_e, P^*_e) + FL(\hat{P}_i, P^*_i),
\end{equation}
where $CLS_e$ and $CLS_i$ represent the emotion and intent classifiers during the pre-training stage.

During the training phase of EI$^2$, we initialize the emotion and intent encoders with pre-trained weights. These encoders are then further updated during the EI$^2$ training.
At last, we adopt a joint training approach and also utilize FL loss as the final loss function $\mathcal{L}_{total}$ for the MC-EIU task.
\begin{equation}
    \mathcal{L}_{total} = FL(\hat{P}_e, P_e) + FL(\hat{P}_e, P_e)
\vspace{-2mm}
\end{equation}

\section{Experiment and Analysis}
\label{sec: Result And Analysis}

\subsection{Baseline}
\label{subsec: Baseline}
\vspace{-2mm}

\begin{wraptable}{r}{0.55 \textwidth}
    \exprimentfontsize
    \centering
    \begin{tabular}{l|cc|cc}
    \hline
    \multirow{2}{*}{\textbf{Systems}} & \multicolumn{2}{c|}{\textbf{English}} & \multicolumn{2}{c}{\textbf{Mandarin}} \\ \cline{2-5} 
                                     & \textbf{Emo}  & \textbf{Int}  & \textbf{Emo}  & \textbf{Int}  \\ \hline
    bc-LSTM                          & 40.11            & 42.81            & 44.74            & 54.76           \\ \hline
    MMIN                             & 40.94            & 43.98            & 46.78           & 56.95           \\ \hline
    MISA                             & 41.46           & 37.64           & 48.51           & 56.26            \\ \hline
    COGMEN                           & \underline{41.70}            & 42.23            & 49.01           & 55.80            \\ \hline
    EmoInt-Trans                     & 40.40            & \underline{44.46}            & \underline{50.47}           & \underline{58.45}            \\ \hline
    \textbf{EI$^2$ (Ours)}                   & \textbf{42.09}   & \textbf{45.53}   & \textbf{55.08}   & \textbf{61.63}   \\ \hline
    $\Delta_{sota}$                  & 0.39   & 1.07   & 4.61   & 3.18   \\ \hline
    \quad w/o History                      & 41.50            & 45.31            & 54.42            & 60.76            \\ \hline
    \quad w/o Interaction                  & 41.44            & 45.25            & 53.77            & 61.03            \\ \hline
    \quad w/o Gating                       & 41.43            & 44.78            & 54.77            & 61.32            \\ \hline
    \quad w/o FL                           & 41.55            & 45.37            & 53.95            & 61.07            \\ \hline
    \quad w/o Pre-training        & 40.48        & 44.99          & 53.57            & 59.62            \\ \hline
    \end{tabular}
    \vspace{-2mm}
    \caption{Comparison results of our EI$^2$, along with all baseline systems and ablation systems, in terms of the WAF metric (\%). 
    The results in \textbf{bold} indicate the highest performance, while the \underline{underlined} results represent the second highest performance.
    The row with $\Delta_{sota}$ means the improvement of EI$^2$ compared to state-of-the-art systems.
}
\label{tab: Performance comparison of our proposed frameworks with existing models}
\end{wraptable}
To validate our MC-EIU datasets and the proposed EI$^{2}$ network, we develop four MC-EIU systems based on state-of-the-art models: 
\textbf{bc-LSTM} \cite{poria2017context} is widely employed in conversational sentiment recognition tasks. It adopts the bi-directional LSTM and multi-head attention mechanism, enabling the model to capture both contextual information and abundant semantic information within utterances. We add additional emotion and intent classifiers to make bc-LSTM support multi-task prediction. 
\textbf{MMIN} \cite{zhao2021missing} incorporates a cascade residual autocoder network and cyclic consistency constraints to learning the robust multimodal joint representation for multimodal emotion recognition. We expand the MMIN model by incorporating an intent classifier, enabling it to recognize emotion and intent simultaneously. 
\textbf{MISA} \cite{hazarika2020misa} introduces a modality-invariant and -specific feature representation learning mechanism to achieve robust multimodal emotion recognition. We also incorporate an additional intent classifier for MISA to support the MC-EIU task. 
\textbf{COGMEN} \cite{joshi2022cogmen} is a multimodal sentiment analysis system based on the contextual Graph Neural Network (GNN) for predicting the sentiment of each speaker per utterance in a conversation. We integrated an extra intent classifier following the feature fusion layer to enable concurrent recognition of sentiment and intent information.
\textbf{EmoInt-Trans} \cite{singh2022emoint} is the state-of-the-art baseline for the MC-EIU task. It adopts MISA \cite{hazarika2020misa} as the backbone, which utilizes adjacent sentences as contextual information.

We choose the Weighted Average F-score (WAF) \cite{poria2019meld} as the evaluation metrics. More implementation details are shown in the Appendix.
\vspace{-2mm}

\subsection{Main Results}
\label{subsec: Main Result}
\vspace{-2mm}
We first compare the recognition performance of our EI$^2$ and the baselines in both English and Mandarin. As shown in Table \ref{tab: Performance comparison of our proposed frameworks with existing models},
we observe that EI$^2$ achieves the highest WAF in both emotion and intent recognition in both languages, clearly outperforming the baseline systems. For example, the WAF of emotion and intent for EI$^2$ in English are 42.09\% and 45.53\%, respectively. In Mandarin, the scores are 55.08\% for emotion and 61.63\% for intent.
These results demonstrate the effectiveness of our method. By learning from multimodal dialog history and employing soft parameter sharing to capture the interaction between emotion and intent, our model achieves impressive joint recognition performance for emotion and intent.
\vspace{-2mm}

\subsection{Ablation Study}
\label{subsec: Ablation Study}
\vspace{-2mm}

We design three different ablation experiments, that are module ablation, task ablation, and modality ablation to further validate the components of EI$^2$ system.

\textbf{Module Ablation}: We conducted ablation experiments on the \textit{multimodal history encoder}, \textit{emotion-intent interaction encoder}, \textit{gate regulator}, and the \textit{pretraining strategy for emotion\&intent encoders}:
1) \textbf{w/o History}: we remove the multimodal history encoder of EI$^2$;
2) \textbf{w/o Interaction}: the features outputted by the emotion and intent encoders are directly used for final prediction;
3) \textbf{w/o Gate}: we remove the gate regulator;
4) \textbf{w/o FL}: we replace Focal Loss referenced in Section \ref{subsec: Training Strategy} with the Cross-Entropy Loss; 5)\textbf{w/o Pre-training}: we randomly initialize the parameters of the Emotion and Intent Encoders during the EI$^2$ training. 

As shown in Table \ref{tab: Performance comparison of our proposed frameworks with existing models}, EI$^2$ outperforms all the ablation models on both datasets.
This provides compelling evidence for the following assertions:
1) The inclusion of conversation history helps analyze the current speaker's emotions and intent states; 
2) By modeling and integrating deep interactive information, the joint understanding task achieves improved performance, and further demonstrates the significance of modeling the interactive information between intent and emotion for joint understanding;
3) After reducing the gating mechanism, the model's performance experienced a significant decline, highlighting the crucial role of the gate regulator in adjusting the weights of binary correlation to the final recognition of emotion or intent;
4) The pre-training of emotion and intent encoders resulted in a notable performance improvement, suggesting their capacity to learn substantial semantic information and distinct features for emotions and intents through pre-training.
\vspace{-2mm}

\begin{wraptable}{l}{0.5 \textwidth}
    \exprimentfontsize
    \centering
    \begin{tabular}{c|cc|cc}
    \hline
    \multirow{2}{*}{\begin{tabular}[c]{@{}c@{}}\textbf{Task  Discription}\end{tabular}} & \multicolumn{2}{c|}{\textbf{English}}      & \multicolumn{2}{c}{\textbf{Mandarin}}      \\ \cline{2-5} 
    & \textbf{Emo}    & \textbf{Int}    & \textbf{Emo}    & \textbf{Int}    \\ \hline
    Single Task                          & 41.50            & 44.91            & 53.74            & 61.59            \\ \cline{1-5} 
    Joint Task                           & \textbf{42.09}   & \textbf{45.53}   & \textbf{55.08}   & \textbf{61.63}   \\ \hline
    \end{tabular}
    \vspace{-2mm}
    \caption{
    Comparison results of \textit{single task} and \textit{joint task}, in terms of the WAF metric  (\%). (The \textbf{bold} numbers have the same meanings as Table 4.)
    }
    \label{tab: Ablation study of task of our frameworks}    
\end{wraptable}
\textbf{Task Ablation}: To investigate the effectiveness of the joint recognition task further, we conduct independent emotion and intent recognition tasks on the MC-EIU dataset.
The results of these separate tasks are presented in Table \ref{tab: Ablation study of task of our frameworks}.
We can observe that the joint task outperforms the recognition results of the single task. 
This further substantiates the strong correlation between intent and emotion and that the joint task can improve prediction accuracy through the deep interaction between emotion and intent.

\textbf{Modality Ablation}: We conduct modality ablation experiments to verify the impact of different modal data, as depicted in Table \ref{tab: Ablation study of modalities of our frameworks}. 
In the unimodal experiments, we find that the textual modality exhibits the highest overall performance, while the visual modality demonstrates the poorest performance. 
This aligns with previous research \cite{fu2022nhfnet}, which concludes that the textual modality contains the most comprehensive semantic information, while the visual modality is comparatively limited in this regard.
Additionally, the model's performance improves as the number of modalities fused increases, with the highest performance achieved when all three modalities are combined. This highlights the effectiveness of integrating complementary information from different modalities.


\section{Conclusion}
\label{Conclusion}

\begin{wraptable}{r}{0.56 \textwidth}
    \exprimentfontsize
    \centering
    \begin{tabular}{c|ccc|cc|cc}
    \hline
    \multirow{2}{*}{\textbf{\begin{tabular}[c]{@{}c@{}}Task\\ Discription\end{tabular}}} & \multicolumn{3}{c|}{\textbf{Modality}} & \multicolumn{2}{c|}{\textbf{English}} & \multicolumn{2}{c}{\textbf{Mandarin}} \\ \cline{2-8} 
    & \textbf{T}  & \textbf{A}  & \textbf{V} & \textbf{Emo}  & \textbf{Int}  & \textbf{Emo}  & \textbf{Int}  \\ \hline
    \multirow{3}{*}{Unimodality}                 & \checked       & -           & -          
    & 38.34            & 44.27           & 49.15            & 58.49            \\ \cline{2-8}
                                                 & -           & \checked       & -          
    & 37.76            & 30.66            & 46.98            & 45.29           \\ \cline{2-8} 
                                                 & -           & -           & \checked      
    & 36.66            & 26.17            & 45.70            & 41.57            \\ \hline 
    \multirow{3}{*}{Bimodality}                  & \checked       & \checked       & -          & 40.50            & 45.35            & 51.68             & 59.90             \\ \cline{2-8} 
                                                 & -           & \checked       & \checked      
    & 39.64             & 30.46             & 49.98             & 45.42             \\ \cline{2-8} 
                                                 & \checked       & -           & \checked      
    & 40.20            & 44.42             & 51.59              & 60.69             \\ \hline 
    Trimodality                                  & \checked       & \checked       & \checked      & \textbf{42.09}   & \textbf{45.53}   & \textbf{55.08}   & \textbf{61.63}   \\ \hline
    \end{tabular}
    \caption{Modality ablation results of the EI$^2$ system, in terms of WAF metric (\%). (The \textbf{bold} numbers have the same meanings as in Table 4.)}
    \label{tab: Ablation study of modalities of our frameworks}
    \vspace{-2mm}
\end{wraptable}
This work introduced a novel dataset called Multimodal Conversational Emotion and Intent Dataset (MC-EIU), which possesses several key properties: annotation diversity (includes emotion and intent labels), modality diversity (encompasses textual, acoustic, and visual data), language diversity (comprises English and Mandarin data), and accessibility.
Furthermore, we proposed an Emotion and Intent Interaction (EI$^2$) Network for the MC-EIU task, which effectively captures the conversational history and the complex interaction between emotion and intent.
Extensive experiments conducted on our dataset demonstrate the effectiveness of our EI$^2$, showcasing its superior performance.
Our work is dedicated to advancing the field of affective computing.

Given the effective modeling of long-range dependencies in conversations by large language models \cite{touvron2023llama}, this characteristic plays a crucial role in understanding the intricate interaction between emotions and intentions, both within individuals and between speakers \cite{deng2023cmcu}. This suggests a promising avenue for future research. Furthermore, previous research has established that emotion stimulus and inertia are significant factors that influence the emotional state of speakers in dialogs \cite{zhao2022m3ed}. However, it remains an open question whether they impact the correlation between emotion and intent.

\bibliographystyle{plainnat}
\bibliography{neurips_data_2024}

\newpage
\appendix



\section{Datasheets for datasets}
\label{sec: Datasheets}

\subsection{Motivation}
\label{Motivation}

Previous works confirm a strong interaction between emotion and intent \cite{welivita2020taxonomy,welivita2020fine,singh2022emoint,deng2023cmcu}.
For instance, \cite{singh2022emoint} emphasizes that particular intents can influence the emotions of the speaker in dialogues.
However, the existing datasets for emotion and intent joint understanding, such as Twitter Customer Support \cite{maharana2022multimodal}, ED \cite{welivita2020taxonomy}, and OSED \cite{welivita2020fine}, only consist of English textual data, which cannot fulfill the requirements for multimodal and multilingual research.
Furthermore, while Singh et al.\cite{singh2022emoint} proposed a multimodal dialogue-based dataset for emotion and intent joint understanding in multimodal conversation, namely EmoInt-MD, it also comprises only English data.
Moreover, it is worth noting that the provided open-source link for EmoInt-MD is currently unavailable. 
As a result, there is no available dataset for emotion and intent joint understanding in the context of multimodal dialogue scenarios.
Without the task-specific datasets, the potential of multimodal emotion and intent joint understanding could not be fully explored, nor deepen the understanding of human complicated affections.

We fill the gap by constructing a large-scale benchmark MC-EIU dataset, which features 7 emotion categories, 9 intent categories, 3 modalities, i.e., textual, acoustic, and visual content, and two languages, i.e., English and Mandarin. Furthermore, it is completely open-source for free access.
We aim for our work to facilitate a deeper exploration of the emotion and intent joint understanding, thereby advancing the field of affective computing.

\subsection{Composition}
\label{subsec: Composition}

\begin{figure}[htbp]
    \centering
    \subfigure[The samples of English audio in MC-EIU]{
    \includegraphics[width=0.95\linewidth]{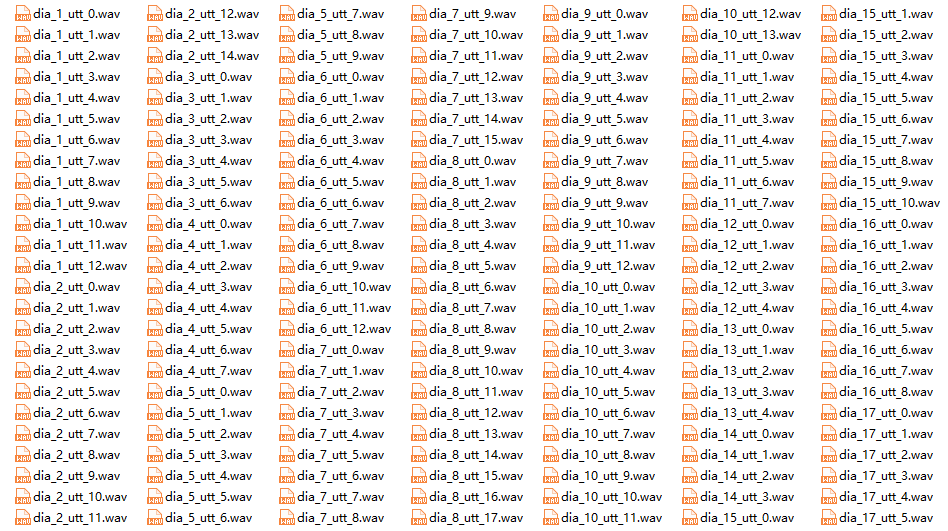}
    }
    \hfill
    \subfigure[The samples of English video in MC-EIU]{
    \includegraphics[width=0.95\linewidth]{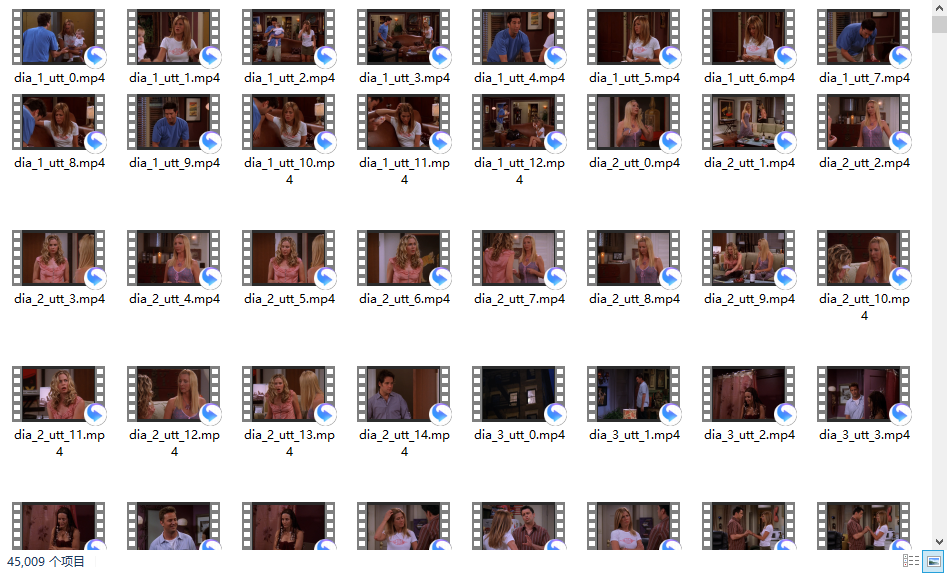}
    }
    \caption{The samples of audio and video files in MC-EIU dataset. All files are named according to a consistent format.}
    \label{fig: Audio and video files}
    \vspace{-2mm}
\end{figure}

The MC-EIU dataset comprises 56,012 utterances, including 4,013 conversations in English and 957 conversations in Mandarin, with a total duration of 53.06 hours.
Each utterance is annotated with emotion, intent, and speaker labels, and contains textual, visual, and acoustic information.
All utterances are recorded in a .CSV file, with each utterance uniquely identified by a Dia\_No and an Utt\_No.
Corresponding video and audio clips for each utterance are stored in .MP4 and .WAV files, respectively.
These files follow a specific naming convention: "dia\_\{Dia\_No\}\_utt\_\{Utt\_No\}.mp4" or "dia\_\{Dia\_No\}\_utt\_\{Utt\_No\}.wav".
The samples of audio and video files in MC-EIU are shown in Figure \ref{fig: Audio and video files}.
We partitioned the MC-EIU dataset into training, validation, and test subsets with proportions of 70\%, 10\%, and 20\%, respectively. In dividing the dataset, we adhered to the following principles:
1) Random assignment of dialogues.
2) Ensuring that all utterances within the same dialogue are assigned to the same subset.
3) Striving to maintain a consistent distribution of labels across each subset.
The statistics of the MC-EIU dataset are presented in Table \ref{tab: MC-EIU Statistic}.

\begin{table}[htbp]
\centering
\exprimentfontsize
\begin{tabular}{lcccccc}
\hline
\multirow{2}{*}{\textbf{Statistics}} & \multicolumn{3}{c}{\textbf{English}}           & \multicolumn{3}{c}{\textbf{Mandarin}}           \\ \cline{2-7} 
                                     & \textbf{Train} & \textbf{Validation} & \textbf{Test} & \textbf{Train} & \textbf{Validation} & \textbf{Test} \\ \hline
\# Modalities                      & (t, a, v)          & (t, a, v)            & (t, a, v)           & (t, a, v)            & (t, a, v)             & (t, a, v)           \\ 
\# Conversations                      & 2,807          & 400            & 806           & 667            & 95             & 195           \\ 
\# Utterances                         & 31,451         & 4,509          & 9,049         & 7,643          & 1,148          & 2,212         \\ 
\# Duration (hours)                   & 28.51          & 4.02           & 8.22          & 8.51           & 1.36           & 2.42          \\ 
Avg.\ Words per Utterance                & 12.68          & 12.49          & 12.76         & 19.11          & 19.91          & 18.14         \\ 
Avg.\ \# of Duration per Utterance (seconds)               & 3.26           & 3.21           & 3.27          & 4.01           & 4.26           & 3.94          \\ 
Avg.\ \# of Utterances per Conversation                         & 11.20          & 11.27          & 11.23         & 11.46          & 12.08          & 11.34         \\ 
Avg.\ \# of Emotions per Conversation                         & 2.58           & 2.57           & 2.60          & 2.41           & 2.54           & 2.42          \\ 
Avg.\ \# of Intents per Conversation                         & 3.29           & 3.86           & 3.87          & 3.18           & 3.24           & 3.10          \\ \hline
\end{tabular}

\caption{Statistic of our MC-EIU. t, a, and v stand for text, acoustic, and visual, respectively.}

\label{tab: MC-EIU Statistic}
\end{table}

\subsection{Collection Process and License}
\label{subsec: Collection Process and License}

\textbf{\textit{Collection Process}}: In this work, we choose 3 famous English TV series and 4 Mandarin TV series as our domain.
Such TV series consist of conversations with utterances in the forms of text, video, and audio, and cover different genres (i.e. family, romance, comedy, lifestyle).
The total number of episodes for English TV series is 716, and for Mandarin TV series, it is 119.
Table \ref{tab: The primary source of conversational data for the MC-EIU dataset.} represents the details of TV series.
In terms of data sources, our resources are more than 1.5 times larger than the resources of M3ED \cite{zhao2022m3ed} and closely 4 times more than those of MELD \cite{poria2019meld}. It covers the majority of emotional dialogue scenarios in real life.
To avoid the inclusion of offensive, discriminatory, or otherwise unethical data in our dataset, we exclude TV series belonging to genres such as horror, thriller, war, crime, etc.
Furthermore, we promptly removed those clips from the dataset if any unethical clips were found in the selected TV series.

\textbf{\textit{License terms}}: To avoid copyright disputes, we ensure that our resources are sourced from publicly accessible platforms.
The MC-EIU dataset is strictly available for research purposes only.
We have designed an appropriate license, specifically the CC BY-NC 4.0, which clearly outlines the proper and responsible usage of the MC-EIU dataset.
It will help guide the user of the MC-EIU dataset in making informed decisions about how the MC-EIU dataset can and cannot be used\footnote{\url{https://github.com/MC-EIU/MC-EIU}}.

\begin{table}[!t]
    \centering
    \exprimentfontsize
    \begin{tabular}{clccl}
    \hline
    \textbf{Language}         & \multicolumn{1}{c}{\textbf{TV series}} & \textbf{\# Sea} & \textbf{\# Epi} & \multicolumn{1}{c}{\textbf{Genres}} \\ \hline
    \multirow{4}{*}{Mandarin} & "\begin{CJK}{UTF8}{gbsn}大江大河\end{CJK}" (The Great River)                         & 1              & 39             & Lifestyle                           \\ 
                              & "\begin{CJK}{UTF8}{gbsn}父母爱情\end{CJK}" (Parental Love)                           & 1              & 44             & Family, Romance                     \\ 
                              & "\begin{CJK}{UTF8}{gbsn}曾少年之小时候\end{CJK}" (The Childhood of Once a Youth)           & 1              & 24             & Family, Romance                     \\  
                              & "\begin{CJK}{UTF8}{gbsn}回家的女儿\end{CJK}" (The Returning Daughter)                  & 1              & 12             & Family                              \\ \hline
    \multirow{3}{*}{English}  & Friends                                 & 10             & 235            & Romance, Comedy                     \\ 
                              & The Big Bang Theory                     & 10             & 231            & Comedy                              \\ 
                              & Modern Family                           & 11             & 250            & Family, Romance                     \\ \hline
    \end{tabular}
    \caption{The details of the data collection for the MC-EIU dataset. `Sea' and `Epi' mean the Season and Episode, respectively}
    \label{tab: The primary source of conversational data for the MC-EIU dataset.}
\end{table}

\subsection{Preprocessing/Cleaning/Labeling}
\label{subsec: Preprocessing}

Please see the Section 3.2 in the main paper.

\subsection{Uses and Distribution}
\label{subsec: Uses and Distribution}
We state that the MC-EIU dataset is suitable for multimodal emotion and intent joint understanding tasks, including emotion recognition, intent recognition, and emotion and intent joint recognition in multimodal conversations.
The copyright of the MC-EIU dataset belongs to the S2 Lab of the College of Computer Science of Inner Mongolia University.
To ensure standardized experimentation and evaluation, we currently release only the feature files for all the data, along with a .CSV file containing the text information and annotations.
Please refer to Appendix \ref{subapp: Feature Extraction} for details on the feature extraction methodology.
Following paper acceptance, the dataset will be made available at \url{https://github.com/MC-EIU/MC-EIU}.

\subsection{Maintenance}
\label{subsec: Maintenance}

Regarding the dataset update iteration, our laboratory will have special personnel to inspect and maintain the MC-EIU dataset every six months.
This includes correcting labeling errors, adding new instances, and deleting outdated instances. 
Meanwhile, we welcome other research teams to use this dataset.

\subsection{Uses}
\label{subsec: Uses}

\begin{itemize}
    \item Has the dataset been used for any tasks already? If so, please provide a description.

    It is proposed to be used for the multimodal emotion and intent joint understanding task.

    \item Is there a repository that links to any or all papers or systems that use the dataset? If so, please provide a link or other access point.

    It is a new dataset. We develop an Emotion and Intent Interaction (EI$^2$) framework as a reference system and release the code at Github.com.

    \item What (other) tasks could the dataset be used for?

    The MC-EIU dataset can also be used for various other tasks such as multimodal emotion recognition, multimodal intent recognition, and dialogue understanding.

    \item Is there anything about the composition of the dataset or the way it was collected and preprocessed/cleaned/labeled that might impact future uses? For example, is there anything that a future user might need to know to avoid uses that could result in unfair treatment of individuals or groups (e.g., stereotyping, quality of service issues) or other undesirable harms (e.g., financial harms, legal risks) If so, please provide a description. Is there anything a future user could do to mitigate these undesirable harms?

    N/A

    \item Are there tasks for which the dataset should not be used? If so, please provide a description.

    N/A

\end{itemize}

\section{Dataset nutrition labels}
\label{sec: Dataset nutrition labels}

\begin{table}[]
\centering
\TaskAblationfontsize
\begin{tabular}{lll}
\cline{1-2}
Name            & Description                                                                                                                                                       &  \\ \cline{1-2}
Subtitle        & The textual content of utterance                                                                                                                                  &  \\
Dia\_No          & The dialogue number to which the current utterance belongs                                                                                                        &  \\
Utt\_No          & The position of utterance within that dialogue                                                                                                                    &  \\
Video\_name      & The TV seies which the utterance is taken                                                                                                                         &  \\
Season          & The season which the utterance is taken                                                                                                                           &  \\
Episode         & The episode which the utterance is taken                                                                                                                          &  \\
Begin\_timestamp & The   starting time of utterance in the TV series                                                                                                                 &  \\
End\_timestamp   & The   ending time of utterance in the TV series                                                                                                                   &  \\
Emotion         & Emotion label (happy, surprise, sad, disgust, anger, fear, neutral)                                                                                               &  \\
Intent          & \begin{tabular}[c]{@{}l@{}}Intent label (questioning, agreeing, acknowledging, sympathizing,\\ encouraging, consoling, suggesting, wishing, neutral)\end{tabular} &  \\
Speaker         & Speaker label (0, 1)                                                                                                                                              &  \\ \cline{1-2}
\end{tabular}
\caption{Dataset Nutrition Labels}
\label{tab: Dataset Nutrition Labels}
\end{table}

Table \ref{tab: Dataset Nutrition Labels} shows the different modules of the MC-EIU dataset nutrition label and their corresponding description.

\section{Data Statements for Natural Language Processing}

We state the MC-EIU dataset from the following aspects:

\begin{itemize}
    \item The full name of the MC-EIU dataset is the \textbf{M}ultimodal \textbf{C}onversational \textbf{E}motion and \textbf{I}ntent Joint \textbf{U}nderstand dataset.
    This dataset addresses the requirements of annotation diversity, modality diversity, language diversity, and accessibility, making it well-suited for the task of joint emotion and intent recognition.
    The copyright belongs to the S2 Lab of the College of Computer Science of Inner Mongolia University.

    \item The dataset is mainly composed of three parts of files (namely .CSV, .MP4, .WAV files), which record text, video, and audio information respectively.
    To ensure standardized experimentation and evaluation, we currently release only the feature files for all the data, along with a .CSV file containing the text information and annotations.

    \item The dataset is mainly manual crawling, which was then processed using a simple Python script we designed to segment all the videos into clips. Subsequently, a manual screening and annotation process was conducted to identify and label the relevant data. Please refer to Section 3 in the main paper for details.

    \item The MC-EIU dataset follows the usage rules and restrictions of the license description dataset, and commercial use, reorganization, and conversion are not allowed. 
    If any organization or individual wants to expand the dataset, permission must be obtained from the copyright owner.
\end{itemize}

\section{Data Accessibility}
\label{sec: Data Accessibility}

The MC-EIU dataset is available at \url{https://github.com/MC-EIU/MC-EIU} and \url{https://pan.baidu.com/s/1gxxr81tVytFTW2UjfTrh-g}.

\section{Accountability frameworks}

\subsection{Data collection and processing specifications}

We declare that our data collection is all in publicly accessible links and follow the following rules in data processing.

\begin{itemize}
    \item Data collection and processing should comply with relevant regulations and ethical requirements, and use appropriate technical tools and methods to ensure data quality and integrity.

    \item Data collection should clarify key information such as data type, source, time, and location.

    \item Data processing should establish a clear data cleaning, conversion, and integration process, and carry out data verification and deduplication.

    \item Data sampling and sample selection should fully consider the impact of research design and sampling error, and carry out statistical inference and reliability analysis.

\end{itemize}

\subsection{Dataset usage and evaluation mechanisms}

Dataset usage and evaluation mechanisms should follow the following principles:

\begin{itemize}
    \item The use of data sets should follow relevant laws and regulations, respect data privacy and intellectual property rights, and prevent abuse and discriminatory results.

    \item Dataset results should be interpreted and applied within a reasonable margin of error, with a full explanation of their limitations and applicability of inferences.

    \item Dataset users should assign corresponding responsibilities and obligations, including data protection, fair use, and social responsibility.

    \item The process of using data sets should be regularly audited and evaluated in order to detect and correct problems in a timely manner and improve the value and credibility of datasets.

\end{itemize}

\section{Author Statement}

On behalf of all the authors, we hereby state that we will assume full responsibility for any violations of rights or issues related to data licensing.

\section{Hosting, Licensing, and Maintenance Plan for MC-EIU Dataset}

Our MC-EIU dataset is a groundbreaking multimodal dataset designed for joint understanding of conversational emotions and intents, which we have created and publicly released. To ensure its accessibility and long-term availability, we have developed a comprehensive plan for hosting, licensing, and maintenance.

\textbf{\textit{Hosting}:} The MC-EIU dataset will be hosted on a reliable and secure server infrastructure, i.e., Github.com and Baidu Drive (refer to Section \ref{sec: Data Accessibility}).
We will ensure fast and uninterrupted access to the dataset for researchers, developers, and interested parties.

\textbf{\textit{Licensing}:} We have designed a proper license (CC BY-NC 4.0) attached to the MC-EIU dataset to clearly describe how to properly and responsibly use the MC-EIU dataset (refer to Section \ref{subsec: Collection Process and License}).

\textbf{\textit{Maintenance}:} We are committed to the continuous maintenance and improvement of the MC-EIU dataset. Regular updates will be provided to address any identified issues and ensure data quality (refer to Section \ref{subsec: Maintenance}).

\newpage

\section{Data Annotation}
\label{app: Data Annotation}

 \subsection{Data Collection}
\label{subapp: Data Collection}

Please refer to Section \ref{subsec: Collection Process and License}.

\subsection{Annotation Guidelines and Annotation Website}
\label{subapp: Annotation Guidelines}

\begin{figure}[htbp]
    \centering
    \includegraphics[width=1.0 \linewidth]{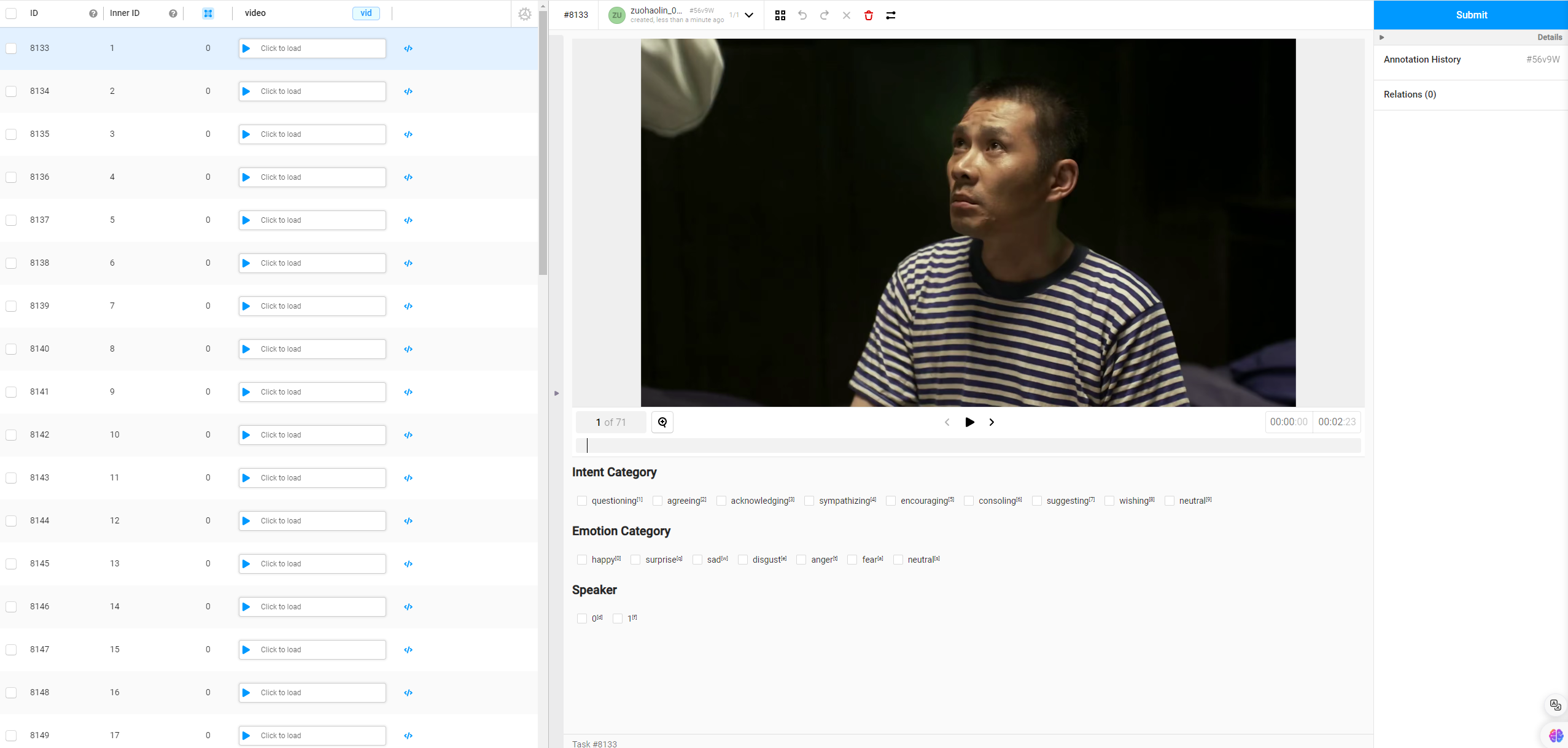}
    \caption{Layout of the annotation platform.}
    \label{fig: annotation_website}
\end{figure}

We presented the relevant details of data annotation in Section 3.2 of the main paper, including the annotation scheme and annotation process.
To facilitate the annotation process for the annotators, we provided a unified data annotation platform, as illustrated in Figure \ref{fig: annotation_website}.

\subsection{Data Format}
\label{subapp: Data Format}
We select some samples from the English and Mandarin datasets of MC-EIU respectively to demonstrate the data format, as shown in Table \ref{tab: Data Format}. 

\setlength{\tabcolsep}{0.0045 \linewidth}{
\begin{table}[htbp]
\tablefontsize
\begin{tabular}{lcccccccccc}
\hline
\multicolumn{1}{c}{\textbf{Subtitle}}                                                                                                                                     & \textbf{Dia\_No} & \textbf{Utt\_No} & \textbf{Video\_name}    & \textbf{\#Sea} & \textbf{\#Epi} & \textbf{Begin\_timestamp} & \textbf{End\_timestamp} & \textbf{Emotion} & \textbf{Intent} & \textbf{Speaker} \\ \hline
Turns out this sweater is made for a woman.                                                 & 34              & 0               & Friends                & 10            & 9             & 00:24:09,900             & 00:24:12,530           & neutral          & neutral         & 0                \\
So why are you still wearing it?                                                            & 34              & 1               & Friends                & 10            & 9             & 00:24:14,910             & 00:24:16,910           & neutral          & questioning     & 1                \\ \hline
\begin{CJK}{UTF8}{gbsn}怎么又是这个呀？\end{CJK}                                    & 519             & 0               & Parental Love       & -             & 4             & 00:34:49,034             & 00:34:50,894           & sad              & questioning     & 0                \\
\begin{CJK}{UTF8}{gbsn}怎么了？这不好啊。\end{CJK}                                   & 519             & 1               & Parental Love       & -             & 4             & 00:34:51,154             & 00:34:52,854           & neutral          & questioning     & 1                \\ \hline
\end{tabular}
\caption{The dataset format of MC-EIU. \textbf{Dia\_No} represents the dialogue number to which the current sentence belongs, while \textbf{Utt\_No} indicates the sentence's position within that dialogue. \textbf{Video\_name}, \textbf{\#Sea}, and \textbf{\#Epi} specify the TV series, season, and episode from which the sentence is taken. \textbf{Begin\_timestamp} and \textbf{End\_timestamp} are used to pinpoint the exact location of the sentence within the episode.}
\label{tab: Data Format}
\end{table}
}

In the MC-EIU dataset, each utterance is uniquely identified by a Dia\_No and an Utt\_No. All of the utterances are stored in .CSV file.
In addition, the corresponding video and audio clips for each utterance are stored as .MP4 and .WAV files, respectively.
These files follow a specific naming format: "dia\_\{Dia\_No\}\_utt\_\{Utt\_No\}.mp4" or "dia\_\{Dia\_No\}\_utt\_\{Utt\_No\}.wav".
The samples of audio and video files in MC-EIU are shown in Figure \ref{fig: Audio and video files}.

\subsection{Feature Extraction}
\label{subapp: Feature Extraction}

We utilize several state-of-the-art pre-trained models to extract the raw features from our datasets.

\begin{wrapfigure}{l}{0.5 \textwidth}
    \centering
    \includegraphics[width=0.55\linewidth]{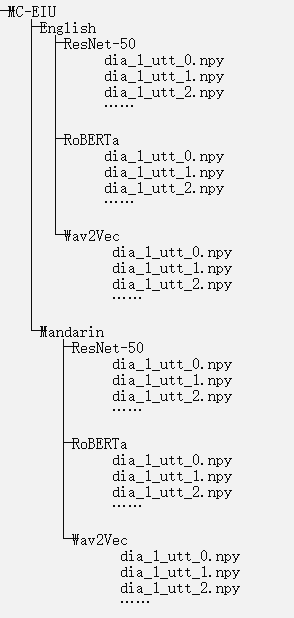}
    \caption{The structure of the feature set.}
    \label{fig: structure of the feature set}
    \vspace{-2mm}
\end{wrapfigure}
\textbf{Textual Features}: 
To extract word-level textual features in English and Chinese, we employ separate RoBERTa \cite{yu2020ch} models \footnote{English: \url{https://huggingface.co/roberta-base}; Chinese: \url{https://huggingface.co/hfl/chinese-roberta-wwm-ext}.} that have been pre-trained on data from each respective language.
The embedding size of the textual features for both languages is 768 dimensions.

\textbf{Acoustic Features}: 
We extract frame-level acoustic features using the Wav2Vec \cite{schneider2019wav2vec} model\footnote{\url{https://github.com/pytorch/fairseq/tree/master/examples/wav2vec}} pre-trained on large-scale Chinese and English audio data. 
The embedding size of the audio features is 512 dimensions.

\textbf{Visual Feature}:
We employ the OpenCV tool to extract scene pictures from each video clip, capturing frames at a 10-frame interval.
Subsequently, we utilize the Resnet-50 \cite{he2016deep} model\footnote{\url{https://huggingface.co/microsoft/resnet-50}} to generate frame-level features for the extracted scene pictures in the video clips.
The embedding size of the video features is 342 dimensions.

All the features are named in the format "dia\_\{Dia\_No\}\_utt\_\{Utt\_No\}.npy".
To differentiate between different modality features, we create separate folders for each modality, naming them according to the corresponding pre-trained model.
All the features are stored within their respective modality folders.
For instance, if we extracted text features using the RoBERTa model, the folder corresponding to the text modality would be named `RoBERTa'.
And all the text features would be stored in the "RoBERTa" folder.
The structure of the feature set is illustrated in Figure \ref{fig: structure of the feature set}.
To ensure standardized experimentation and evaluation, we currently release only the feature files for all the data, along with a .CSV file containing the text information and annotations.

\subsection{Category Distribution}
\label{subapp: Category Distribution}

\begin{figure}[]
    \centering
    \subfigure[Emotion distribution of Mandarin (left) and English (right).]{
    \includegraphics[width=0.95\linewidth]{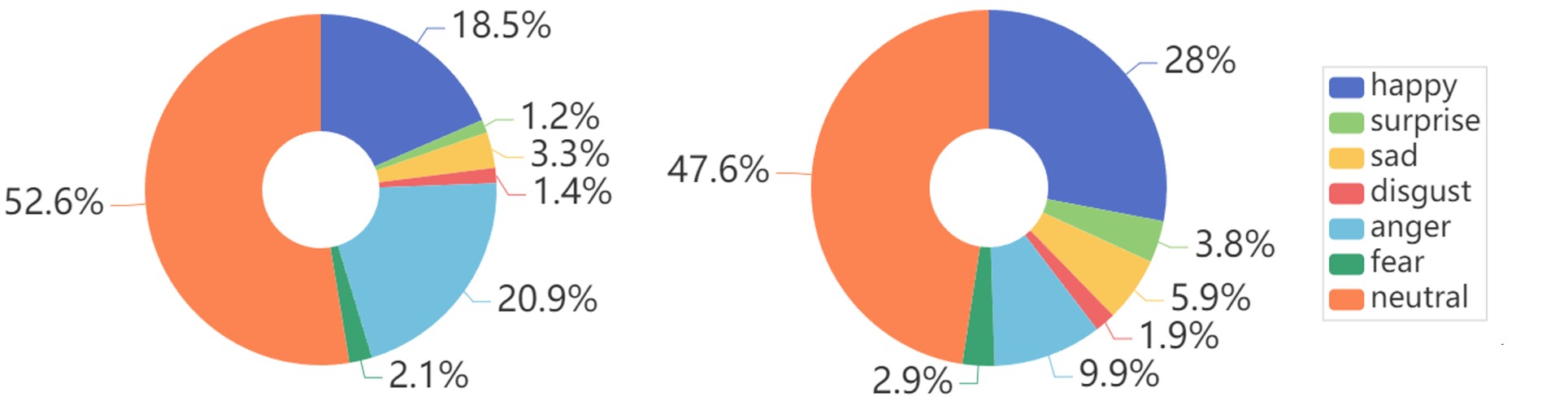}
    }
    \hfill
    \subfigure[Intent distribution of Mandarin (left) and English (right)]{
    \includegraphics[width=0.95\linewidth]{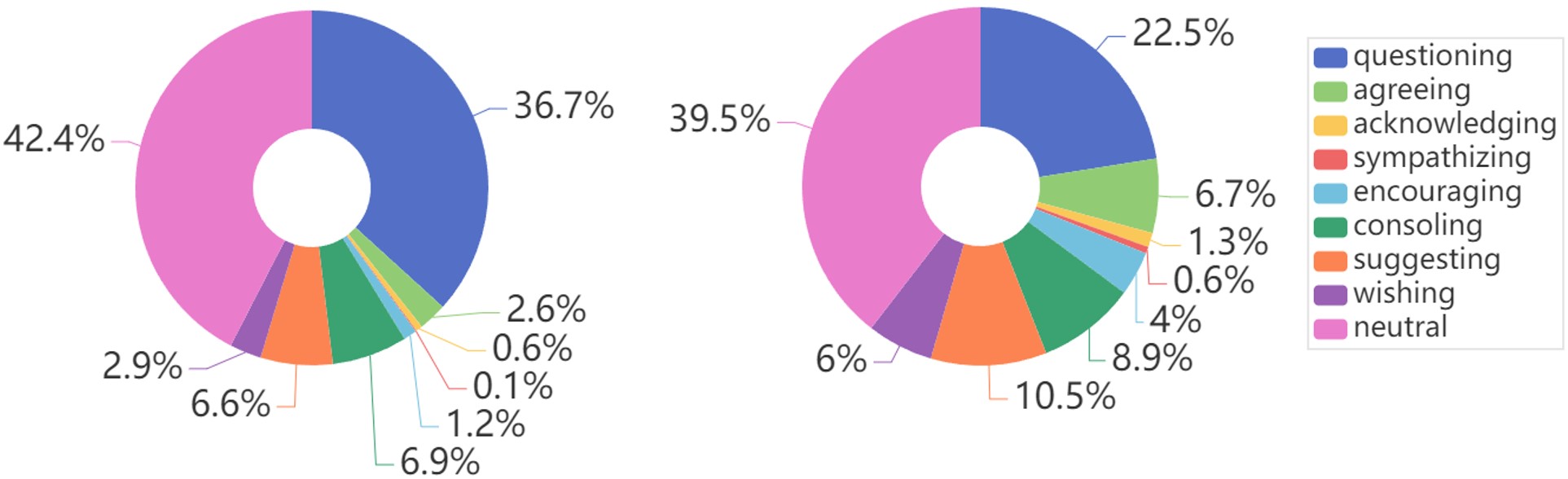}
    }
    \caption{The data distribution of the MC-EIU dataset.}
    \label{fig: distribution of categories}
    \vspace{-2mm}
\end{figure}

We show the distribution of emotion and intent categories for the MC-EIU dataset in Figure \ref{fig: distribution of categories}.
Similar to M3ED, MELD, and IEMOCAP, our dataset also exhibits category imbalance.
In terms of emotion labels, the top three categories in our dataset are \textit{neutral}, \textit{happy}, and \textit{anger}, while \textit{disgust} and \textit{fear} are relatively less represented, similar to the distribution in MELD.
Regarding intent labels, the most abundant categories are \textit{neutral}, \textit{questioning}, and \textit{suggesting}, while \textit{sympathizing} is the least represented, aligning with the OSED dataset.
We present a detailed distribution of the data across the train, validation, and test sets in Table \ref{tab: The data distribution}.

The presence of category imbalance in our dataset can be attributed to the following reasons:
\begin{itemize}
    \item \textbf{Deliberate maintenance of category imbalance:} Although researchers tend to use balanced datasets in experiments to explore model performance, category imbalance is a more common and challenging scenario in the real world \cite{japkowicz2002class, he2009learning, fernandez2018smote, johnson2019survey}. For instance, in work environments, individuals may conceal their emotions during interactions. In daily life, a stable and harmonious social environment promotes the expression of positive emotions (such as happiness and joy) while reducing the expression of negative emotions (such as anger and sadness). Therefore, intentionally maintaining this imbalance helps better simulate and address real-world problems, enhancing the applicability and generalization of the models.

    \item \textbf{Bias During Data Collection:} As described in Section \ref{subsec: Collection Process and License}, we deliberately excluded TV shows such as thrillers and crime dramas to avoid aggressive or unethical content.
    However, in genres like family dramas, comedies, and soap operas, conflicts and contradictions among characters are less common.
    Consequently, it becomes challenging to obtain data related to certain negative emotions, such as disgust and fear.
    Additionally, the categories of intent can be influenced by the emotion categories.
    For instance, when one party in a conversation displays fear, the other party typically expresses sympathy and encouragement. If there is a relative scarcity of fear-related data, there will also be a corresponding scarcity of data related to sympathy and encouragement. This bias in data collection contributes to the category imbalance observed in the dataset.
\end{itemize}

\begin{table}[]
\begin{tabular}{c@{\hspace{5pt}}lc@{\hspace{12pt}}c@{\hspace{12pt}}c@{\hspace{12pt}}c@{\hspace{12pt}}c@{\hspace{12pt}}c}
\hline
\multicolumn{2}{c}{\multirow{2}{*}{Annotation}} & \multicolumn{3}{c}{English} & \multicolumn{3}{c}{Mandarin} \\
\multicolumn{2}{c}{}                            & Train    & Validation   & Test   & Train    & Validation    & Test   \\ \hline
\multirow{7}{*}{Emotion}     & happy            & 8810     & 1286    & 2526   & 1447     & 189      & 400    \\
                             & surprise         & 1242     & 158     & 332    & 89       & 21       & 27     \\
                             & sad              & 1739     & 251     & 660    & 243      & 31       & 87     \\
                             & disgust          & 607      & 108     & 129    & 116      & 10       & 28     \\
                             & anger            & 3146     & 418     & 912    & 1608     & 242      & 453    \\
                             & fear             & 869      & 114     & 273    & 184      & 20       & 18     \\
                             & neutral          & 15038    & 2174    & 4217   & 3956     & 635      & 1199   \\ \hline
\multirow{9}{*}{Intent}      & questioning      & 7133     & 963     & 2035   & 2804     & 385      & 844    \\
                             & agreeing         & 2102     & 293     & 603    & 199      & 37       & 54     \\
                             & acknowledging    & 410      & 50      & 124    & 43       & 10       & 15     \\
                             & sympathizing     & 188      & 28      & 49     & 5        & 1        & 1      \\
                             & encouraging      & 1227     & 177     & 388    & 85       & 23       & 22     \\
                             & consoling        & 2841     & 383     & 785    & 535      & 78       & 147    \\
                             & suggesting       & 3270     & 516     & 962    & 519      & 82       & 128    \\
                             & wishing          & 1881     & 297     & 536    & 236      & 25       & 60     \\
                             & neutral          & 12399    & 1802    & 356    & 3217     & 507      & 941    \\ \hline
\end{tabular}
\caption{The statistic of data distribution across the train, validation, and test sets.}
\label{tab: The data distribution}
\end{table}


\subsection{Ethical Considerations}
\label{subapp: Ethical Considerations}
\vspace{-2mm}

In data selection, we ensured that our resources came from public platforms to avoid copyright disputes.
Additionally, we adopted the CC BY-NC 4.0 license to restrict the dataset's usage to non-commercial purposes only.
Furthermore, to avoid incorporating offensive or unethical content, we imposed limitations on selecting dataset resources, excluding TV shows of genres such as crime and thriller.
Please refer to Section \ref{subsec: Collection Process and License} for more details.

In terms of employee compensation, our pricing model is based on the quantity of annotated data, with a payment of 0.1 CNY per annotated utterance.
As shown in Table \ref{tab: MC-EIU Statistic}, the average duration per utterance of our data is approximately 3.7 seconds.
Assuming each utterance requires watching the video twice for accurate annotation, annotating one utterance would take a maximum of 10 seconds.
Therefore, an annotator can annotate 360 utterances per hour, equivalent to an hourly wage of 36 CNY.
This wage level is significantly higher than that of typical part-time jobs in our city.
To ensure accuracy, we use a majority voting method for final annotations, meaning each utterance undergoes three rounds of annotation. Thus, the total annotation cost is calculated as follows:
56012 (total number of utterances) * 3 * 0.1 CNY (cost per utterance) = 16803.6 CNY.

\begin{wrapfigure}{r}{0.58 \textwidth}
    \centering
    \subfigure[EmoInt-Trans]{
    \includegraphics[width=0.45\linewidth]{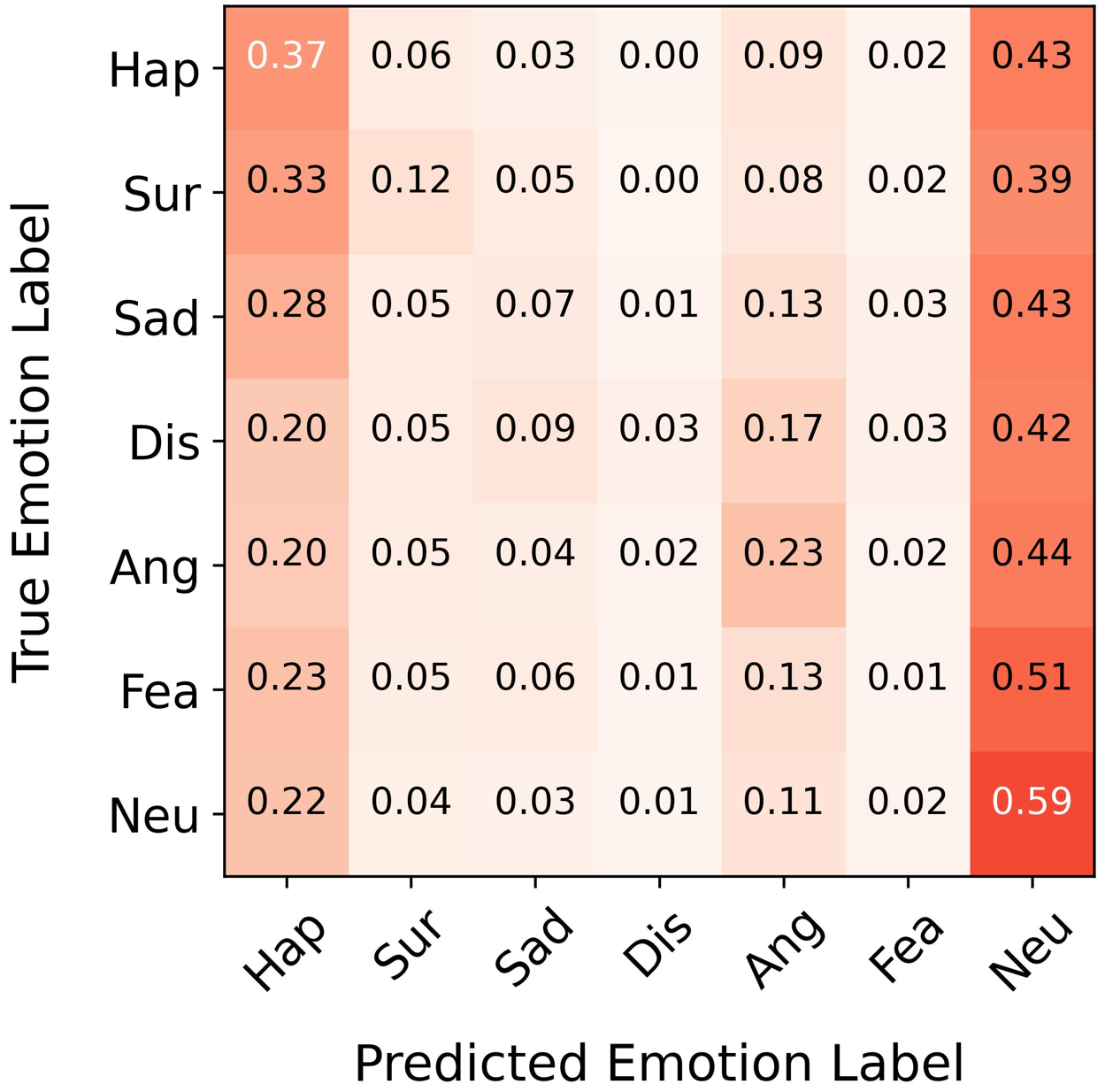}
    }
    \hfill
    \subfigure[EI$^2$]{
    \includegraphics[width=0.45\linewidth]{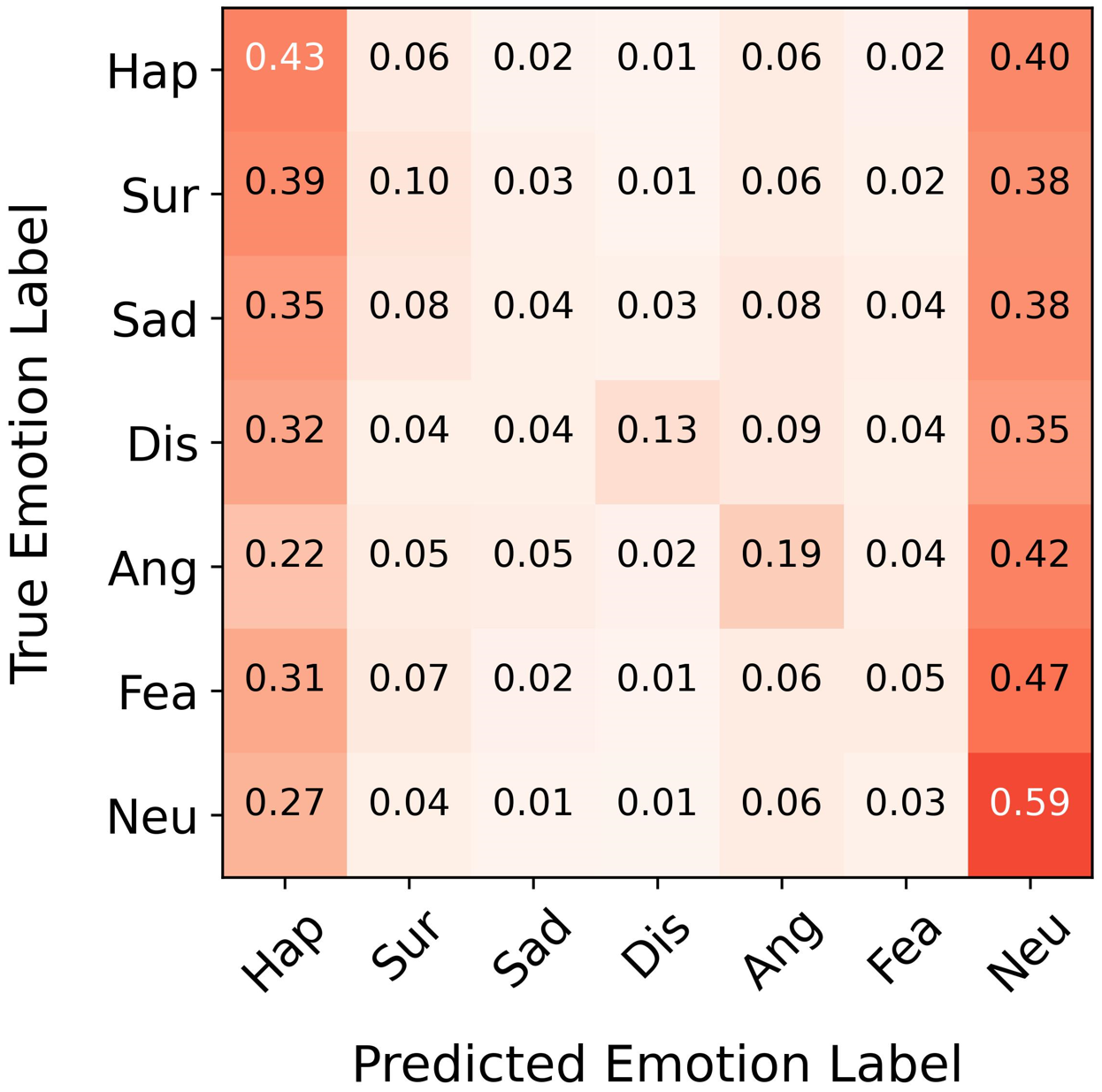}
    }
    \hfill
    \subfigure[EmoInt-Trans]{
    \includegraphics[width=0.45\linewidth]{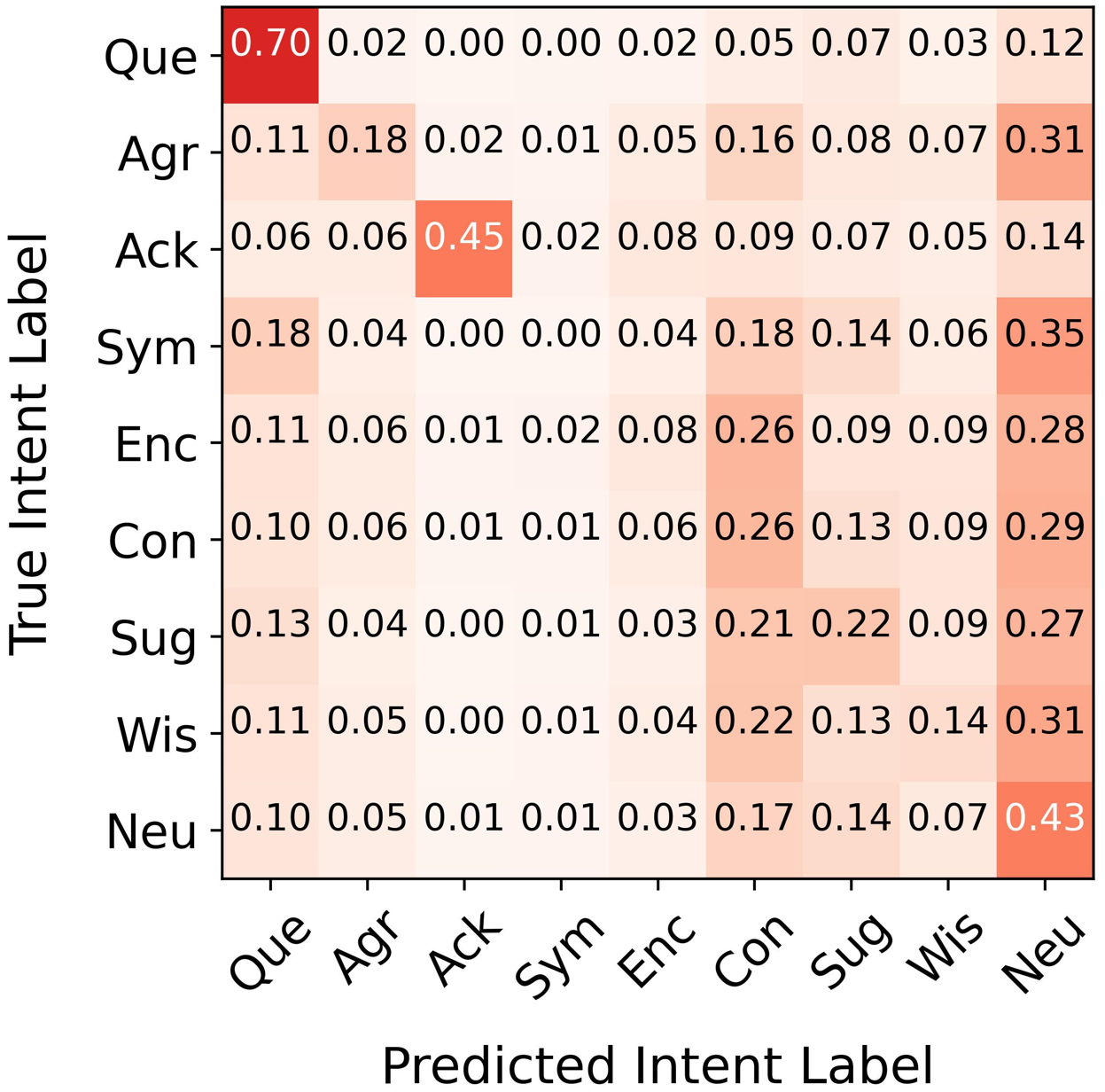}
    }
    \hfill
    \subfigure[EI$^2$]{
    \includegraphics[width=0.45\linewidth]{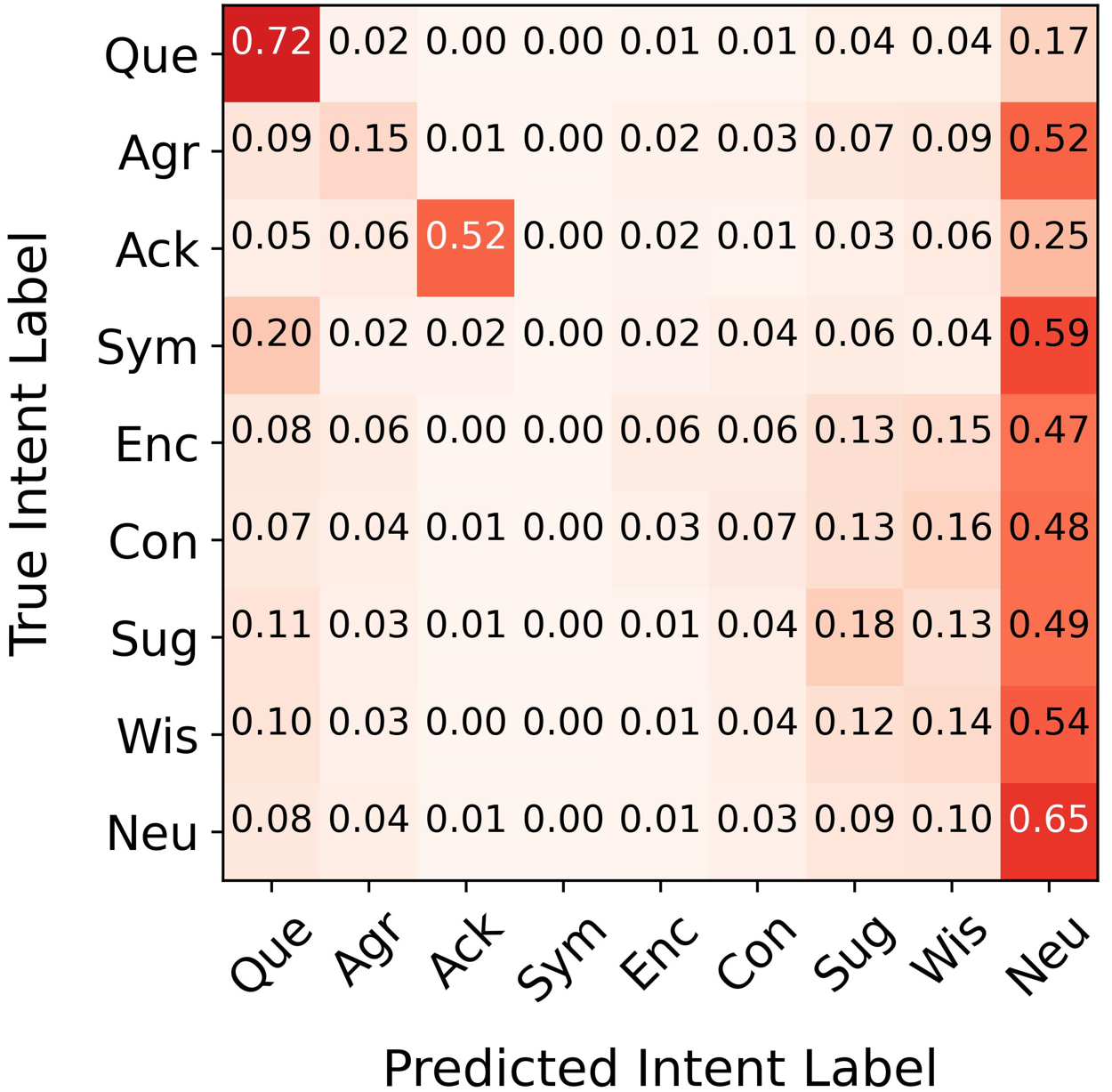}
    }
    \caption{Confusion matrices of emotion and intent recognition results for the EmoInt-Trans and EI$^2$ models on the MC-EIU English dataset. Specifically, (a) and (b) represent the confusion matrices for emotion recognition, whereas (c) and (d) correspond to the confusion matrices for intent recognition. Each cell represents the ratio of samples from each true category that are predicted as the corresponding category.
    }
    \label{fig: Intent recognition confusion matrix}
\end{wrapfigure}
In a nutshell, it is important to note that the dataset being studied in this paper does not involve any ethical concerns.
\vspace{-2mm}

\section{Extra Experiment Results and Analysis}
\label{app: Extra Experiment}
\vspace{-2mm}

\subsection{Implementation Details}
\label{subapp: Implementation Details}

Our proposed framework is implemented using PyTorch. 
The hidden size of $F^v$, $F^a$, and $F^t$ is 128.
We set the attention head number for the Transformer Network, Cross Attention, and Fusion Attention as 4.
We select the Adam optimizer \cite{DBLP:journals/corr/KingmaB14} and initialize the learning rate to 0.0002.
We dynamically update the learning rate using the Lambda LR \cite{wu2020deep} approach.
The batch size is 32, and the epoch of both pre-training and training phases is set to 60.
To ensure the objectivity of the experimental results, we conducted all experiments three times and reported the average results.
All experiments were performed on a single NVIDIA A100 graphics card.
\vspace{-2mm}

\subsection{Category Imbalance Issue}
\label{subapp: Category Imbalance Issue}

As mentioned in Appendix \ref{subapp: Category Distribution}, our MC-EIU dataset, like many other datasets, suffers from category imbalance issues.
Taking the advanced model EmoInt-Trans for the MC-EIU task as an example, we compare our model with its predictions on more detailed categories. 
In our experiments on the English data, we display the detailed F1 scores for the predictions of all emotion and intent categories in the confusion matrix of Figure \ref{fig: Intent recognition confusion matrix}.

In the upper part of Figure \ref{fig: Intent recognition confusion matrix}, we can find that both models face difficulties in recognizing \textit{fear}, \textit{sad}, and \textit{disgust}, which can be attributed to the insufficiency of samples for these categories. 
Similar observations can be made in Figure \ref{fig: Intent recognition confusion matrix}(c), (d). These two models encounter difficulties in recognizing the categories of \textit{sympathizing} and \textit{encouraging}, as evident from Figure \ref{fig: Intent recognition confusion matrix}(c) and (d). This is further supported by Figure \ref{fig: distribution of categories}, which illustrates that these two categories have the lowest number of samples.

These observations serve as evidence that the category imbalance issue poses a significant challenge for both models. Moving forward, we will actively explore effective solutions to tackle this challenge. Specifically, this includes the following aspects:

(1) \textbf{Introducing effective constraints:} In the design of our framework, we have already incorporated Focal Loss to mitigate the issue of data imbalance (as discussed in Section 4.3 of the main paper). Ablation experiments have shown that the introduction of Focal Loss improves the model's performance, indicating that the use of specific constraint methods can alleviate category imbalance. However, Focal Loss is a relatively basic method and may not be highly effective when there are significant differences in the number of instances per category. Therefore, we will further explore more suitable constraints to alleviate the category imbalance issue

(2) \textbf{Data augmentation:} As mentioned in Section \ref{subsec: Collection Process and License}, we avoided TV series of genres such as thrillers, horror, and crime when selecting the data.
However, these genres often contain richer negative emotions and corresponding intents compared to genres like romantic or family dramas.
By avoiding these types of TV shows, we inadvertently created a scarcity of data related to negative emotions.
To address the category imbalance issue, we will supplement the dataset with new data during the maintenance process.
It is important to note that this does not mean we aim to transform the dataset into a perfectly balanced one.
We intend to maintain the presence of the category imbalance in the dataset while controlling the disparities between different categories.


\subsection{Case Study}
\label{subapp: Case Study}

In this section, we continue to use the EmoInt-Trans system as the baseline for comparison. To further validate the effectiveness of our EI$^2$ method, we present the case study.

\begin{wrapfigure}{r}{0.58 \textwidth}
    \centering
    \includegraphics[width=0.7\linewidth]{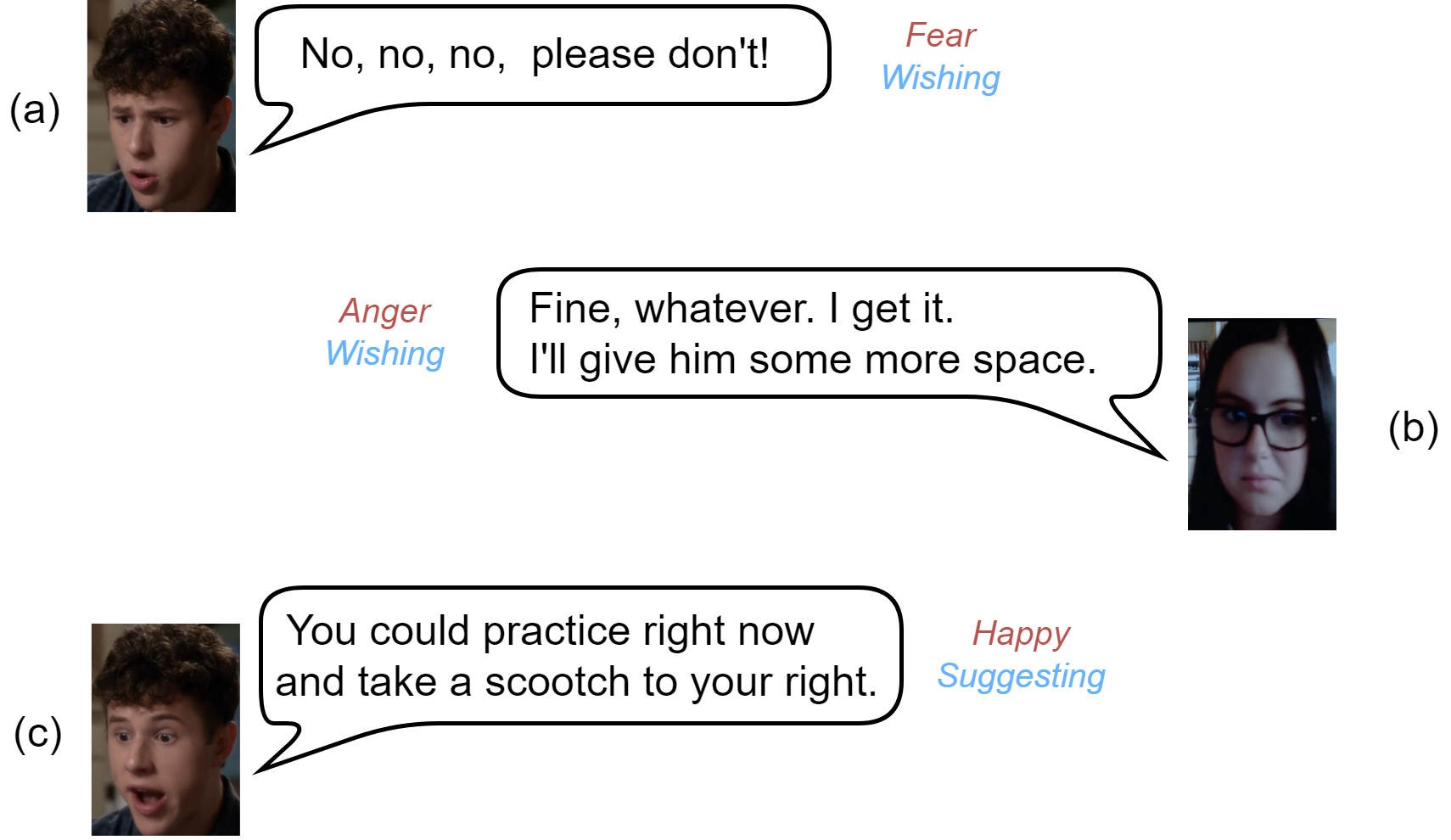}
    \caption{A conversational sample, that consists of 3 utterances, from the MC-EIU dataset. The red text means the emotion label, while the blue text means the intent label.}
    \label{fig: Example of a conversational scenario from the MC-EIU dataset}
    \vspace{-2mm}
\end{wrapfigure}
Figure \ref{fig: Example of a conversational scenario from the MC-EIU dataset} presents a conversational sample, that consists of 3 utterances, from the MC-EIU dataset. We adopt the EmoInt-Trans and EI$^2$ models to predict emotion and intent labels, respectively.
The Ground Truth labels and the prediction results are shown in Table \ref{tab: The prediction performance of EmoInt-Trans and EII}.

We can observe that the predictions by EI$^2$ are consistently correct, whereas the results of EmoInt-Trans are not.
Taking utterance (a) as an example, EmoInt-Trans fails to make accurate predictions due to its limited ability to model deep interactions between emotion and intent.
As shown in Figure 1 in the main paper, the proportion of ``Fea-Wis" is smaller compared to ``Neu-Wis", indicating a weaker correlation between \textit{fear} and \textit{wishing}.
Therefore, in the joint recognition process, the impact of interaction between \textit{fear} and \textit{wishing} on the final prediction is relatively low. 
EmoInt-Trans cannot dynamically adjust the impact of interaction information on the final prediction, leading to incorrect predictions. Different from EmoInt-Trans, our model incorporates the \textit{Emotion-Intent Interaction Encoder}, which enables deep interactions between emotion and intent to be modeled while controlling the weight of interaction information during final prediction. As a result, our model accurately predicts the emotion and intent labels of this example.

\setlength{\tabcolsep}{0.012 \linewidth}{
\begin{table}[htbp]
    \centering
    \exprimentfontsize
\begin{tabular}{c|cc|cccc}
\hline
\multirow{3}{*}{\textbf{Sample ID}} & \multicolumn{2}{c|}{\multirow{2}{*}{\textbf{Ground Truth}}} & \multicolumn{4}{c}{\textbf{Predicted Labels}}                                           \\ \cline{4-7} 
                                  & \multicolumn{2}{c|}{}                                       & \multicolumn{2}{c|}{\textbf{EmoInt-Trans}}       & \multicolumn{2}{c}{\textbf{EI$^2$}} \\ \cline{2-7} 
                                  & \textbf{Emo}                 & \textbf{Int}                 & \textbf{Emo} & \multicolumn{1}{c|}{\textbf{Int}} & \textbf{Emo}     & \textbf{Int}     \\ \hline
(a)                               & Fear                         & Wishing                      & \textcolor{red}{\textbf{Neutral}}      & \multicolumn{1}{c|}{\textcolor{green}{Wishing}}      & \textcolor{green}{Fear}             & \textcolor{green}{Wishing}          \\ \hline
(b)                               & Anger                        & Wishing                      & \textcolor{green}{Anger}        & \multicolumn{1}{c|}{\textcolor{red}{Neutral}}      & \textcolor{green}{Anger}            & \textcolor{green}{Wishing}          \\ \hline
(c)                               & Happy                        & Suggesting                   & \textcolor{red}{\textbf{Neutral}}      & \multicolumn{1}{c|}{\textcolor{green}{Suggesting}}   & \textcolor{green}{Happy}            & \textcolor{green}{Suggesting}       \\ \hline
\end{tabular}
\caption{The prediction performance of EmoInt-Trans and EI$^2$ for the sample shown in Figure \ref{fig: Example of a conversational scenario from the MC-EIU dataset}. The red text indicates the wrong results, and the green text means the right results.}
\label{tab: The prediction performance of EmoInt-Trans and EII}
\end{table}
}
As for the incorrect prediction of utterance (c) by EmoInt-Trans, it is due to its limited ability to model contextual information beyond adjacent sentences.
Since the request ``\textit{please don't}" in utterance (a) is fulfilled in utterance (b) (``\textit{Fine, whatever. I get it.}"), utterance (c) exhibits a \textit{happy} emotion.
However, EmoInt-Trans fails to capture the distant historical information, resulting in the inability to recognize the reason for the emotional transition and leading to incorrect predictions.

\section{Impact}
\label{app: Impact}

\textbf{Positive Impacts:}
\begin{itemize}
    \item \textbf{Innovation:} As the first open-source dataset in the emotion and intent joint understanding field, The MC-EIU dataset provides a new resource that can spur innovation and development in emotion and intent recognition. Furthermore, this dataset enables researchers to explore the intricate interactions between emotion and intent, ultimately advancing the development of emotion recognition tasks.

    \item \textbf{Collaboration:} The availability of the MC-EIU dataset encourages collaboration among researchers across different institutions and industries, fostering a more unified and rapid advancement in the field.

    \item \textbf{Improved Systems:} The dataset can help improve the accuracy and reliability of systems that rely on emotion and intent recognition, such as virtual assistants, customer service bots, and interactive entertainment systems.
\end{itemize}

\textbf{Negative Impacts:}
\begin{itemize}
    \item \textbf{Category Imbalance:} Due to the category imbalance phenomenon in the MC-EIU dataset, it is possible for models adapted to category balance to inaccurately recognize minority categories, leading to misjudgments in model performance. 
    Therefore, we encourage more researchers to pay attention to the category imbalance issue. 
\end{itemize}

\section{Limitation}
\label{app: Limitation}

(1) Our proposed dataset does not reach the requirement of "balance".
As described in Appendix \ref{subapp: Category Distribution} and \ref{subapp: Category Imbalance Issue}, our dataset exhibits category imbalance.
This is a common issue in real-world scenarios and a challenge that researchers need to address.
In our future work, we will actively explore methods to tackle the category imbalance issue and encourage future researchers to pay attention to such issues as well.

(2) The framework we established is a basic model.
The main focus of this paper is the dataset release, with the baseline model serving as an accompanying system.
The baseline system has emphasized the interaction between emotion and intent, which is crucial for multimodal emotion-intent joint recognition, even though the ideas behind the baseline may seem simple.
In the future, we will further explore new methods to delve deeper into the complex relationship between emotion and intent.

(3) We do not explore the performance of the large language model on the emotion and intent joint understanding task.
Given the effective modeling of long-range dependencies in conversations by large language models (LLMs) \cite{touvron2023llama}, this characteristic plays a crucial role in understanding the intricate interaction between emotions and intentions, both within individuals and between speakers \cite{deng2023cmcu}.
We have already discussed the possibility of using the LLMs for emotion and intent joint understanding in Section 6 of the main paper.
While we attempted to fine-tune large-scale multimodal models for this task, we faced challenges during training due to limited computational resources.
In the future, we will seek opportunities to collaborate with other laboratories to conduct joint research on using LLMs for emotion and intent recognition. 
Additionally, we hope this direction becomes a new research hotspot, attracting more researchers to contribute to this field.

\end{document}